\documentclass[11pt]{article}

\usepackage[final]{acl}
\usepackage{times}
\usepackage{latexsym}

\usepackage[T1]{fontenc}
\usepackage[utf8]{inputenc}

\usepackage{microtype}

\usepackage{inconsolata}

\usepackage{graphicx}

\usepackage{amsmath}
\usepackage{booktabs}
\usepackage{multirow}
\usepackage{array}
\usepackage{geometry}
\usepackage{adjustbox}
\usepackage{xcolor}
\usepackage{xspace}
\usepackage{diagbox}
\usepackage{xcolor, colortbl} % 用于背景颜色
\usepackage{arydshln}

\usepackage{amsfonts}
\usepackage{algorithm}
\usepackage{algorithmic}
\usepackage{xcolor} % 用于显示代码中的灰色注释
\usepackage{amsmath}
\usepackage{pgfplots}
\usepackage{wrapfig}
\usepackage{makecell}

\title{HEALing Entropy Collapse: Enhancing Exploration in Few-Shot RLVR via Hybrid-Domain Entropy Dynamics Alignment}

\author{
 \textbf{Zhanyu Liu\textsuperscript{1,4}\thanks{Equal contribution.}},
 \textbf{Qingguo Hu\textsuperscript{1,4}\footnotemark[1]\thanks{Project lead.}},
 \textbf{Ante Wang\textsuperscript{2}},
 \textbf{Chenqing Liu\textsuperscript{1,4}},
 \textbf{Zhishang Xiang\textsuperscript{1,4}},
 \\
 \textbf{Hui Li\textsuperscript{1,4}},
 \textbf{Delai Qiu\textsuperscript{3}},
 \textbf{Jinsong Su \textsuperscript{1,4}\thanks{Corresponding author.}}
 \\
 \textsuperscript{1}School of Informatics, Xiamen University
 \\
 \textsuperscript{2}Tsinghua University
 \textsuperscript{3}Xiamen Unisound Intelligence Technology Co., Ltd
 \\
 \textsuperscript{4}Key Laboratory of Digital Protection and Intelligent Processing of Intangible Cultural
 \\
 Heritage of Fujian and Taiwan (Xiamen University), Ministry of Culture and Tourism, China
 \\
 \texttt{\{liuzhanyu, huqingguo\}@stu.xmu.edu.cn, jssu@xmu.edu.cn}
}
\begin{document}
\maketitle

\begin{abstract}

Reinforcement Learning with Verifiable Reward (RLVR) has proven effective for training reasoning-oriented large language models, but existing methods largely assume high-resource settings with abundant training data. In low-resource scenarios, RLVR is prone to more severe entropy collapse, which substantially limits exploration and degrades reasoning performance.
To address this issue, we propose \textbf{H}ybrid-domain \textbf{E}ntropy dynamics \textbf{AL}ignment (\textbf{HEAL}), a framework tailored for few-shot RLVR. HEAL first selectively incorporates high-value general-domain data to promote more diverse exploration. Then, we introduce Entropy Dynamics Alignment (EDA), a reward mechanism that aligns trajectory-level entropy dynamics between the target and general domains, capturing both entropy magnitude and fine-grained variation.
Through this alignment, EDA not only further mitigates entropy collapse but also encourages the policy to acquire more diverse exploration behaviors from the general domain.
Experiments across multiple domains show that HEAL consistently improves few-shot RLVR performance.
Notably, using only 32 target-domain samples, HEAL matches or even surpasses full-shot RLVR trained with 1K target-domain samples.
Our code is available at \url{https://github.com/XMUDeepLIT/HEAL}. 
\end{abstract}
\section{Introduction}

\begin{figure}[t]
	\centering  
	\includegraphics[width=\linewidth]{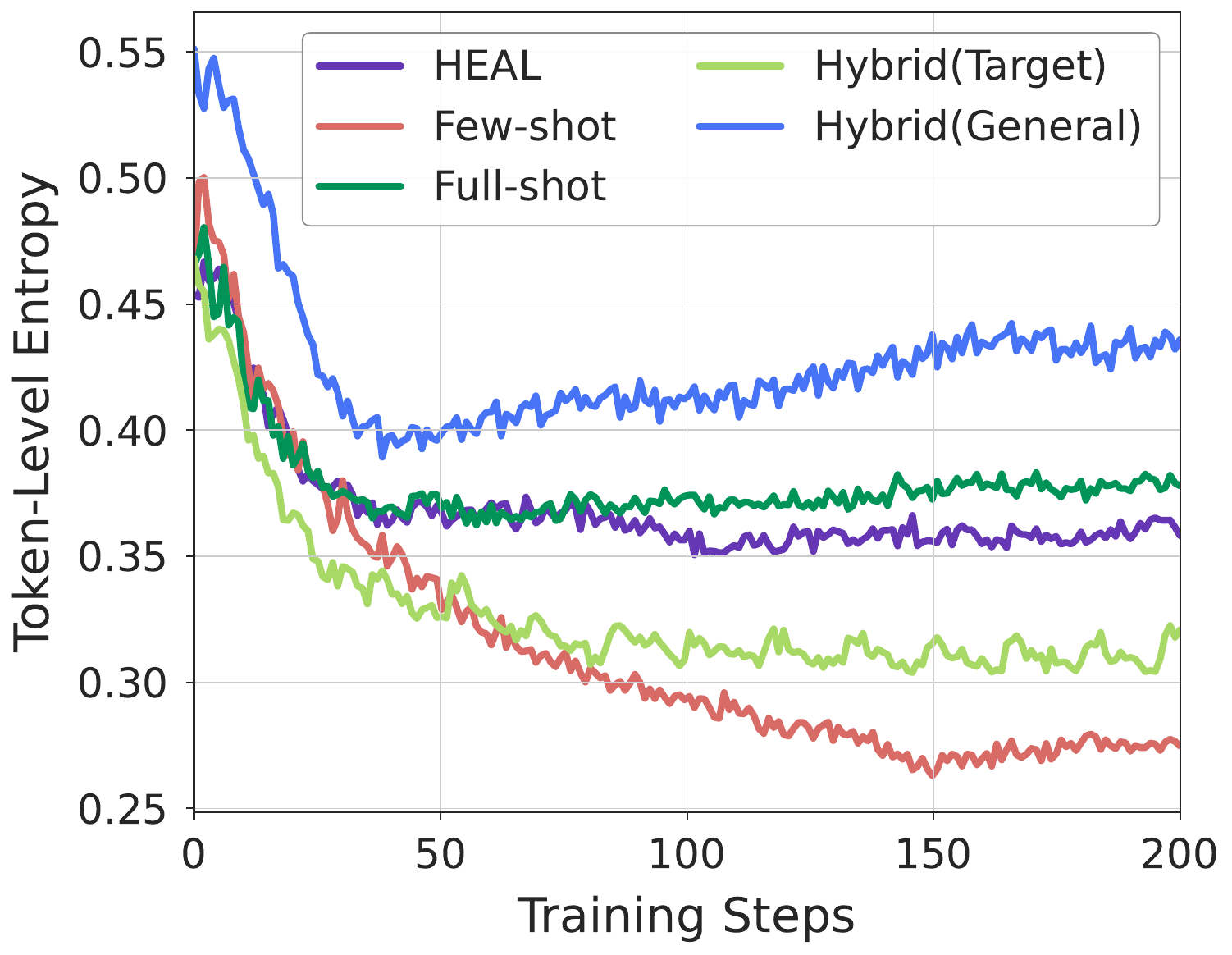}  
	\caption{Average token-level entropy during training under different settings. Few-shot and Full-shot denote RLVR trained with few-shot and sufficient target-domain data, respectively. Hybrid(Target) and Hybrid(General) indicate few-shot RLVR augmented with randomly sampled general-domain data, evaluated on target- and general-domain data. Notably, our framework enables few-shot RLVR to achieve entropy magnitude comparable to full-shot training.}
	\label{fig:preliminary_entropy_curve}  
\end{figure}

Recent breakthroughs in Large Language Models (LLMs)~\cite{liu2024deepseek,yang2025qwen3,hu2025sigma} have given rise to a new generation of reasoning-oriented systems, exemplified by models such as OpenAI o1~\cite{openai2024learning} and DeepSeek-R1~\cite{deepseek2025deepseek}, which demonstrate substantially improved performance on complex reasoning tasks.
A key technique in achieving such success is Reinforcement Learning with Verifiable Reward (RLVR)~\cite{lambert2024tulu,deepseek2025deepseek}, which utilizes rule-based outcome rewards to provide a binary feedback on the correctness of a model's final answer. This simple yet effective mechanism not only mitigates reward hacking~\cite{cui2025primerewardhack}, but also eliminates the need to train complex reward models~\cite{schulman2017ppo}.

Despite its demonstrated effectiveness, existing RLVR research~\cite{shao2024deepseekmath, yu2025dapo, yue2025vapo, liu2025specrl} has predominantly focused on high-resource
domains, where large volumes of high-quality training samples are readily available. However, this assumption of abundant and reliable reward signals does not hold in many real-world domains, such as medical reasoning~\cite{zhang2025medlowres} and other specialized knowledge domains~\cite{guha2023legalbench, zhang2025phybench}, where training data for RLVR are often scarce.
In this context, the policy is prone to rapid overfitting to a few generated trajectories, which prematurely limits exploration and leads to more severe entropy collapse compared to high-resource scenarios~\cite{yue2025does, cui2025entropy}.
As shown in Figure~\ref{fig:preliminary_entropy_curve}, compared to sufficient training data (Full-shot), Few-shot RLVR exhibits a significantly lower entropy magnitude.
While recent studies~\cite{he2025skywork,zheng2025gspo} have proposed various methods to mitigate entropy collapse in RLVR, they often overlook the scarcity of training data, and directly applying them under such conditions may be sub-optimal.

In this paper, we propose \textbf{H}ybrid-domain \textbf{E}ntropy dynamics \textbf{AL}ignment (\textbf{HEAL}), a novel framework specifically designed to boost exploration diversity for RLVR under low-resource scenarios. Our proposed framework is built upon two key components:

We firstly incorporate readily available general-domain data.
Intuitively, although general-domain data may not provide domain-specific knowledge, it offers fundamental reasoning patterns and thus encourages more diverse exploration. Much like a human learner applying general skills to a new domain~\cite{GICK1980306humanlearning}, this hybrid training prevents the policy from prematurely narrowing its search space, thereby notably mitigating entropy collapse in the target domain. To further avoid incurring excessive training costs from general-domain data, we adopt a data selection strategy that retains only a small set of high-value samples based on their reasoning uncertainty and exploratory diversity.

Despite the benefits of this hybrid training, we find that the policy’s entropy in the target domain remains substantially lower than that of the general domain.
To bridge this gap, we then introduce a novel reward mechanism, termed Entropy Dynamics Alignment (EDA). Unlike conventional approaches that naively increase entropy without constraints~\cite{wang2025reinforcement, liu2025prorl}, which can limit policy exploitation or even destabilize training, EDA leverages general-domain data as a reference to guide the policy toward more diverse exploration in target domain. Specifically, EDA constructs \textit{trajectory-level entropy dynamics} for both target- and general-domain data, which captures not only the token-level entropy magnitude but also fine-grained variation. By comparing entropy dynamics within and across domains, EDA rewards trajectories that exhibit stronger inter-domain similarity, thereby effectively encouraging alignment between target- and general-domain entropy characteristics (magnitude and variation). Through this mechanism, the policy not only elevates entropy in the target domain in a controlled manner but also acquires more diverse exploratory behaviors guided by the general domain.

To demonstrate the effectiveness of our proposed framework, we conduct extensive experiments across multiple domains, including Medicine, Physics, Code, and Math. Experiments on the Qwen3~\cite{yang2025qwen3} and LLaMA-3.2~\cite{dubey2024llama} series of models show that our framework consistently and substantially improves the performance of few-shot RLVR. Remarkably, with only 32 target-domain samples, our approach matches or even surpasses full-shot RLVR trained with 1K target-domain samples. Further analyses indicate that our framework also outperforms existing entropy regularization methods in low-resource scenarios.

\begin{figure*}[!t]  
	\centering  
	\includegraphics[width=\textwidth]{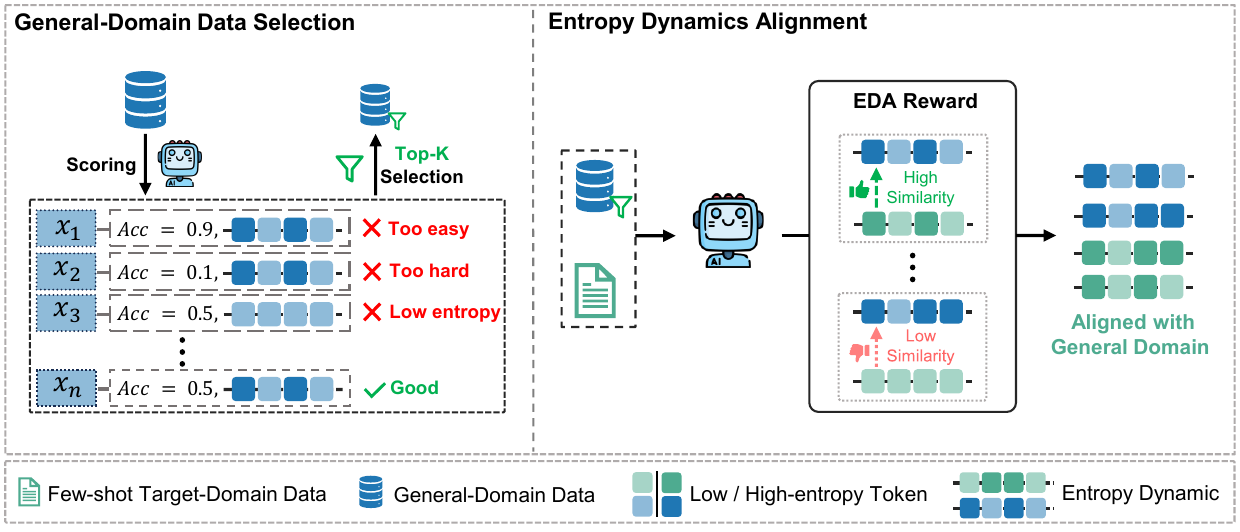}  
	\caption{
        Overview of our proposed HEAL framework. \textit{Left}: We incorporate a small set of high-value general-domain data into few-shot RLVR to promote diverse exploration and mitigate entropy collapse in the target domain. \textit{Right}: Entropy Dynamics Alignment reward guides the policy by aligning the trajectory-level entropy dynamics of few-shot target-domain data with those of the selected general-domain data, thereby further encouraging controlled increases in entropy and more diverse exploratory behaviors.
    }
	\label{fig:overviewofmethod_new}  
\end{figure*}

\section{Related Work}

\paragraph{Low-Resource RLVR}
Extensive research has explored Supervised Fine-Tuning (SFT)~\cite{chen2023alpagasus,ivison2025large} and Reinforcement Learning from Human Feedback (RLHF)~\cite{muldrew2024active,liu2024enabling,das2025active} as primary paradigms for improving the performance of LLMs under low-resource settings. Despite their success, how to effectively apply RLVR in such scenarios remains an open problem.
To address this challenge, emergent paradigms leverage self-play and autonomous feedback~\cite{huang2025r,zhao2025absolute,hu2025boosting} to enhance reasoning capabilities without human annotations. However, due to the lack of reliable supervision signals, such self-improvement is often constrained by the model’s internal knowledge boundaries, thereby limiting out-of-domain generalization.
Recent data-centric studies~\cite{li2025limr} have demonstrated the potential of leveraging a very small amount of data to boost the reasoning ability of LLMs (i.e., few-shot RLVR). Nevertheless, exploration of this paradigm remains preliminary, and its associated challenges have yet to be fully addressed.
For instance, \citet{wang2025reinforcement} reports severe training collapse when applying 1-shot RLVR under certain settings, without identifying the underlying mechanisms. To bridge this gap, our work reveals the challenges of entropy collapse in few-shot RLVR and proposes effective solutions to mitigate this issue.

\paragraph{Entropy Perspectives in RLVR}
Recent RLVR research~\cite{yue2025does,dang2025weight, min2026dhpo} highlights the role of entropy in the exploration–exploitation trade-off, showing that higher entropy promotes diverse reasoning trajectories, whereas lower entropy facilitates convergence. However, in RLVR, entropy often exhibits excessive contraction, causing the policy to lose exploratory capacity, which is referred to as entropy collapse~\cite{cui2025entropy,he2025skywork,liu2025prorl,wu2025quantile}.
To mitigate entropy collapse during RLVR, existing studies~\cite{sheng2024hybridflow,he2025skywork,wang2025beyond,cheng2025reasoning,chen2025seed} have explored a range of regularization-based strategies to preserve exploratory diversity. Nevertheless, these works typically do not account for the effect of data scale. In contrast, we show that entropy collapse becomes substantially more severe under low-resource scenarios.
To address this issue, our proposed HEAL introduces hybrid-domain training and guides the policy to align fine-grained entropy characteristics across domains, thereby substantially mitigating entropy collapse and promoting diverse exploration.
\section{Our Proposed Framework}
In this section, we present \textbf{H}ybrid-domain \textbf{E}ntropy dynamics \textbf{AL}ignment (\textbf{HEAL}), a novel framework designed to mitigate entropy collapse in RLVR under low-resource scenarios. As illustrated in Figure~\ref{fig:overviewofmethod_new}, HEAL consists of two key components:
We first incorporate carefully selected general-domain data into the training process. %to 
By doing so, we can effectively introduce more diverse exploratory behaviors beyond the target domain, preventing the policy from prematurely collapsing its search space (\S \ref{sec:mix-general}).
To further promote exploration in the target domain, we then introduce Entropy Dynamics Alignment (EDA), which guides the policy to align the entropy characteristics between the target and general domains (\S \ref{sec:reward}).

\subsection{Hybrid Training with Selected General-Domain Data}
\label{sec:mix-general}

When encountering a new domain, human learners often demonstrate remarkable transfer learning capabilities by leveraging general skills and problem-solving strategies~\cite{GICK1980306humanlearning}, even with limited prior knowledge.
Building on this insight, we incorporate readily available general-domain data (e.g., commonsense reasoning) into the few-shot RLVR training process. Although these samples lack domain-specific knowledge, they provide fundamental reasoning patterns and thus promote more diverse exploration.

To validate this intuition, we randomly selected 1K samples from a commonsense-reasoning dataset~\cite{talmor-etal-2019-commonsenseqa} and directly combined them with few-shot target-domain data for RLVR training.
As illustrated in Figure~\ref{fig:preliminary_entropy_curve}, this simple mixture notably alleviates entropy collapse in the target domain and leads to notable performance improvements compared to training only on few-shot target-domain data.

However, indiscriminately incorporating excessive general-domain data would be prohibitively expensive in terms of computational cost. Therefore, to maximize the efficacy of this hybrid training, we select only a small set of high-value general-domain samples based on two complementary criteria: \emph{Reasoning Uncertainty} and \emph{Exploratory Diversity}.

\begin{itemize}
    \item \textit{Reasoning Uncertainty}: 
    Given $N$ generated trajectories $\{y_1, \dots, y_N\}$ for an input question $x$, we first compute the average accuracy of these trajectories as $ \mathrm{Acc}(x) = \frac{1}{N} \sum_{i=1}^{N} \mathbb{I}(y_i = y_{\text{gt}})$, 
    where $\mathbb{I}(\cdot)$ denotes the indicator function and $y_{\text{gt}}$ is the ground-truth answer.
    We then define the reasoning uncertainty as $\mathrm{Uncertainty}(x) = 1 - 2  \left| \mathrm{Acc}(x) - \frac{1}{2} \right|$.
    Samples with high uncertainty are preferred, as they have been shown to be the most effective for enhancing reasoning capabilities~\cite{huang2025r}.

    \item \textit{Exploratory Diversity}: 
    To select samples that elicit more diverse exploration, we collect the top 20\% of tokens with the highest entropy from each generated trajectory and compute their average entropy as the exploratory diversity, denoted as $\mathrm{Diversity}(x)$.
    A higher exploratory diversity score indicates that the sample induces more varied exploratory behaviors~\cite{wang2025beyond}.
\end{itemize}

Finally, we combine the two criteria into a single scalar score:
\begin{equation}
c(x) = \mathrm{Uncertainty}(x) \cdot \mathrm{Diversity}(x).
\end{equation}
We select the top-$K$ general-domain samples with the highest composite scores to serve as high-quality training data.

\subsection{Entropy Dynamics Alignment}
\label{sec:reward}

While hybrid training helps reduce entropy collapse in the target domain, we observe that the policy’s entropy in the target domain remains considerably lower than that in the general domain, as shown in Figure~\ref{fig:preliminary_entropy_curve}. This gap indicates that merely mixing general-domain data is insufficient to mitigate entropy collapse in the target domain. To bridge this gap, we propose Entropy Dynamics Alignment (EDA), a novel reward mechanism designed to guide the policy toward more diverse exploration in the target domain by aligning its entropy characteristics with those observed in the general domain.

\paragraph{Trajectory-Level Entropy Dynamics}
We first provide the definition of trajectory-level entropy dynamics.
Formally, given a trajectory $y_i$, we define its entropy dynamics as the sequence of token-level entropies over generation steps, $\tau_y = (\mathcal{H}_1, \mathcal{H}_2, \dots, \mathcal{H}_{\lvert y_i \rvert
})$, where $H_t$ is the entropy over the vocabulary distribution at timestep $t$, conditioned on the input and previously generated tokens.
This sequence captures not only the token-level entropy magnitude of a trajectory, but also its fine-grained variation, thereby providing a richer representation than direct aggregation~\cite{he2025skywork, cheng2025reasoning}.

The discrepancy between the entropy dynamics of two trajectories can be measured using a suitable similarity function $s(\cdot, \cdot)$. Specifically, given two trajectories $y_1$ and $y_2$, the discrepancy between their entropy dynamics is defined as $s(\tau_{y_1}, \tau_{y_2})$.
Since trajectories may vary in length, we first apply an interpolation-based alignment strategy to normalize their lengths before computing the similarity. Implementation details are provided in the Appendix~\ref{appendix:distance_func}.

\paragraph{Entropy Dynamics Alignment Reward}
\label{subsec:trajectory_transfer}

The core idea of this reward is to align the entropy characteristics between target-domain and general-domain data, in terms of both token-level magnitude and fine-grained variation.
During hybrid training, we collect entropy dynamics from both the target and general domains, denoted as $\mathcal{B}_{\text{tgt}}$ and $\mathcal{B}_{\text{gen}}$, respectively. For a target-domain entropy dynamics $\tau_{y_i} \in \mathcal{B}_{\text{tgt}}$, we compute two kinds of similarity measures.

\begin{itemize}
    \item \textit{Intra-Domain Similarity}: 
    We define the intra-domain similarity as the maximum similarity between $\tau_i$ and other entropy dynamics from the same domain:
    \begin{equation}
        \mathcal{S}_{\text{intra}}(\tau_{y_i})
        =
        \max_{\tau_{y_j} \in \mathcal{B}_{\text{tgt}},\, j \neq i}
        s(\tau_{y_i}, \tau_{y_j}).
    \end{equation}
    % which quantifies how closely the entropy dynamics of $\tau_i$ resemble those of its target-domain peers.

    \item \textit{Inter-Domain Similarity}:
    Similarly, we define the inter-domain similarity as the maximum similarity between $\tau_i$ and entropy dynamics from the general domain:
    \begin{equation}
        \mathcal{S}_{\text{inter}}(\tau_{y_i})
        =
        \max_{\tau_{y_k} \in \mathcal{B}_{\text{gen}}}
        s(\tau_{y_i}, \tau_{y_k}).
    \end{equation}
    % which measures how well the entropy dynamics $\tau_i$ aligns with those observed in the general domain.
\end{itemize}

We then assign a reward to trajectories whose entropy dynamics exhibit higher inter-domain similarity than intra-domain similarity. These trajectories are considered strong exemplars for promoting alignment across domains. Based on this, we define the EDA reward for the trajectory $y_i$ as a binary signal:
\begin{equation}
    r_{\text{EDA}}(y_i)
    =
    \begin{cases}
        1, & \text{if } \mathcal{S}_{\text{inter}}(\tau_{y_i}) > \mathcal{S}_{\text{intra}}(\tau_{y_i}), \\
        0, & \text{otherwise}.
    \end{cases}
\end{equation}

By doing so, this reward mechanism not only elevates entropy in the target domain in a controlled manner but also promotes fine-grained alignment of entropy variation, allowing the policy to implicitly acquire more diverse exploratory behaviors from the general domain.

\paragraph{Policy Update}
Standard RLVR uses a deterministic accuracy reward, denoted as $r_{\text{Acc}}(y_i)$, which assesses whether the final answer of trajectory $y_i$ matches its ground truth. In our framework, we combine this accuracy reward with our proposed EDA reward. Formally, the final reward for trajectory $y_i$ is defined as $r(y_i)=r_{\text{Acc}}(y_i) + r_{\text{EDA}}(y_i)$.
Consequently, the policy optimization objective is formulated as minimizing the following loss function:
\begin{equation}
    \mathcal{L}_{\text{RLVR}}(\theta) = -\mathbb{E}_{{x} \sim \mathcal{D}, {y} \sim \pi_{\theta}(\cdot|{x})} \left[r({y}) \right]
\end{equation}
where $\theta$ represents the policy parameters to be optimized, $\mathcal{D}$ denotes the training dataset containing prompts $x$, $\pi_{\theta}(\cdot|x)$ is the likelihood of the generated trajectory $y$.

\definecolor{TechBlue}{RGB}{0, 123, 255}
\begin{table*}[!t]
\small
% 调整行高，让表格不拥挤（替代 addlinespace）
\renewcommand{\arraystretch}{1.125} 
% 关键设置：消除 booktabs 对竖线的截断，使竖线贯穿全表
\setlength{\aboverulesep}{0pt}
\setlength{\belowrulesep}{0pt}
\setlength{\tabcolsep}{3.85pt}
\centering

% 列定义：增加了竖线 | 分隔 DataSize, Medicine, Physics, Code
\begin{tabular}{l | c @{\hspace{3pt}} c | c c c | c c c | c c c}
\toprule
& \multicolumn{2}{c|}{Data Size} & \multicolumn{3}{c|}{Medicine} & \multicolumn{3}{c|}{Physics} & \multicolumn{3}{c}{Code} \\
\cmidrule(lr){2-3} \cmidrule(lr){4-6} \cmidrule(lr){7-9} \cmidrule(lr){10-12}
% 注意：最后一列 Code 移到了最右边
 &$ \left| \mathcal{D}_{tgt} \right| $  & $\left|\mathcal{D}_{gen}\right|$ & MBul. & MedXQA & Avg. & C-Eval & WebIns. & Avg. & HEval+  & LCBv5 & Avg. \\
\midrule
Qwen3-1.7B-Base                                  &          &           & 41.78 & 31.06 & 36.42 & 49.53 & 5.69 & 27.61 & 55.49              & 5.39              & 30.44 \\
\textcolor{gray}{$\hookrightarrow$} Full-shot         & 1K &   0       & \underline{49.01} & \textbf{40.45} & \textbf{44.73} & \underline{60.68} & \underline{11.79} & \underline{36.24} & \underline{60.98}  & \underline{17.37} & \underline{39.18} \\
\textcolor{gray}{$\hookrightarrow$} Few-shot          & 32       &    0      & 45.07 & 38.33 & 41.70 & 57.27 & 7.32 & 32.30 & 56.71              & 12.57             & 34.64 \\
\textcolor{gray}{$\hookrightarrow$} Only-General &  0       &10K   & 42.11 & 32.73 & 37.42 & 55.39 & 7.72 & 31.56 & 58.54              & 10.18             & 34.36 \\
\textcolor{gray}{$\hookrightarrow$} Hybrid & 32       &   384      & 45.39 & 38.64 & 42.02 & 59.74 & 8.13 & 33.94 & 57.93              & 13.77             & 35.85 \\
% 科技蓝背景，!10 代表极高透明度(仅10%颜色浓度)
\rowcolor{TechBlue!10}
\textcolor{gray}{$\hookrightarrow$} \textbf{HEAL}         & 32       &   384      & \textbf{49.34} & \underline{40.00} & \underline{44.67} & \textbf{62.19} & \textbf{12.20} & \textbf{37.20} & \textbf{62.80}     & \textbf{19.76}    & \textbf{41.28} \\
\hdashline
Qwen3-4B-Base                                    &        &            & 43.09 & 33.93 & 38.51 & 53.88 & 8.94 & 31.41 & 66.46             & 16.77             & 41.62 \\
\textcolor{gray}{$\hookrightarrow$}  Full-shot         & 1K &    0       & \textbf{68.09} & \textbf{45.30} & \textbf{56.70} & 74.10 & \textbf{15.04} & \underline{44.57} & \underline{78.05} & 23.95             & \underline{51.00} \\
\textcolor{gray}{$\hookrightarrow$} Few-shot          & 32       &     0      & 55.26 & 43.18 & 49.22 & 73.35 & 11.79 & 42.57 & 71.95             & 22.75             & 47.35 \\
\textcolor{gray}{$\hookrightarrow$} Only-General & 0& 10K  & 49.67 & 39.70 & 44.69 & 68.62 & 8.13 & 38.38 & 70.12             & 20.35             & 45.24 \\
\textcolor{gray}{$\hookrightarrow$} Hybrid & 32   &    384     & 55.92 & 42.73 & 49.33 & \underline{76.75} & 11.38 & 44.07 & 74.39             & \underline{24.55} & 49.47 \\
% 科技蓝背景
\rowcolor{TechBlue!10}
\textcolor{gray}{$\hookrightarrow$} \textbf{HEAL}         & 32   &    384     & \underline{57.24}  & \underline{44.39} & \underline{50.82} & \textbf{80.15} & \underline{13.82} & \textbf{46.99} & \textbf{79.88}    & \textbf{26.95}    & \textbf{53.42} \\
\hdashline
LLaMA3.2-3B-Instruct                             &        &            & 40.46 & 26.82 & 33.64 & 33.46 & 4.88 & 19.17 & 48.17             & 11.38             & 29.78 \\
\textcolor{gray}{$\hookrightarrow$}  Full-shot         & 1K &    0       & \textbf{84.87} & \textbf{42.88} & \textbf{63.88} & \underline{36.67} & \underline{6.91} & \underline{21.79} & 51.22             & \underline{16.17} & \underline{33.70} \\
\textcolor{gray}{$\hookrightarrow$} Few-shot          & 32       &     0      & 56.91 & 31.36 & 44.14 & 35.54 & 6.10 & 20.82 & 50.61             & 12.57             & 31.59 \\
\textcolor{gray}{$\hookrightarrow$} Only-General & 0& 10K  & 48.68 & 28.48 & 38.58 & 36.11 & 5.69 & 20.90 & 51.22             & 13.77             & 32.50 \\
\textcolor{gray}{$\hookrightarrow$} Hybrid & 32   &    384     & 58.55 & 30.15 & 44.35 & 35.16 & 5.28 & 20.22 & \underline{51.83} & 15.57             & 33.70 \\
% 科技蓝背景
\rowcolor{TechBlue!10}
\textcolor{gray}{$\hookrightarrow$} \textbf{HEAL}         & 32   &    384     & \underline{60.53} & \underline{32.27} & \underline{46.40} & \textbf{37.05} & \textbf{7.72} & \textbf{22.39} & \textbf{52.44}    & \textbf{16.77}    & \textbf{34.61} \\
\bottomrule
\end{tabular}
\caption{Comprehensive results on reasoning benchmarks across three target-domains. $\left| \mathcal{D}_{tgt} \right|$ and $\left| \mathcal{D}_{gen} \right|$ denote the sizes of the target- and general-domain training data, respectively. The best results are highlighted in bold, and the second-best results are underlined.}
\label{tab:code_medicine_physics}
% \vspace{-0.25cm}
\end{table*}

\section{Experiments}
\label{sec:experiments}

\subsection{Setup}
\label{subsec:setup}

\paragraph{Datasets} 

We evaluate our framework across three domains: \emph{Medicine}, \emph{Physics}, and \emph{Code}. These domains have relatively scarce open-source and high-quality data, particularly in Medicine and Physics. 
% For each domain, we first collect $1\text{K}$ samples to construct a full-shot dataset.
For the Medicine domain, we utilize MedBullets~\cite{chen-etal-2025-benchmarking} and MedXpertQA~\cite{zuo2025medxpertqa}, following the data-splitting pipeline proposed by~\citet{qiu2025openmed} to construct the training and test sets.
For the Physics domain, we use WebInstruct~\cite{ma2025generalreasoner} for training.
For the Code domain, we employ LiveCodeBench (v1--v4)~\cite{jain2024livecodebench} as the training dataset.
To simulate real-world RLVR under low-resource scenarios, we randomly sample a small number of training samples from each domain to construct few-shot datasets.
This setup enables an effective evaluation of our framework under few-shot settings and facilitates direct comparison with the high-resource scenarios.

For the general-domain dataset, we adopt CommonsenseQA~\cite{talmor-etal-2019-commonsenseqa} for hybrid training.
CommonsenseQA is a multiple-choice benchmark designed to evaluate models’ commonsense reasoning over everyday scenarios, encompassing background world knowledge, causal relations, and social reasoning.
Therefore, it exhibits minimal overlap with the knowledge required in the aforementioned target domains.

\paragraph{Models and Training Details}
To evaluate the generalizability of our framework, we perform RLVR across multiple models with different architectures, including Qwen3-1.7B-Base, Qwen3-4B-Base~\cite{yang2025qwen3}, and LLaMA-3.2-3B-Instruct~\cite{dubey2024llama}.
We adopt VERL~\cite{sheng2024hybridflow} as the RLVR training pipeline.
For all models, we sample 8 trajectories per input. The rollout temperature is set to 0.7 for Qwen3 and 0.6 for LLaMA-3.2, following~\citet{wang2025reinforcement}.
The training batch size and mini-batch size are both set to 128, with a maximum input length of 1024 tokens and a maximum trajectory length of 3072 tokens.
For the remaining hyperparameters, we follow the training recipe of GRPO~\cite{shao2024deepseekmath}.
All experiments are conducted on 8 × A100 (80 GB) GPUs.

\paragraph{Evaluation Benchmarks}
For the Medicine domain, evaluation is conducted on the test splits of MedBullets (Mbul.) and MedXpertQA (MedXQA).
For the Physics domain, we evaluate on the physics subsets of both C-Eval~\cite{huang2023ceval} and WebInstruct (WebIns.)~\cite{ma2025generalreasoner}.
For both the Medicine and Physics domains, we adopt the Qwen2.5-Math~\cite{yang2024qwen2} evaluation pipeline and report the Avg@4 score.
For the Code domain, we evaluate on LiveCodeBench~v5 (LCBv5)~\cite{jain2024livecodebench} and HumanEval Plus (HEval+)~\cite{chen2021codex}, using the official LiveCodeBench~\cite{jain2024livecodebench} evaluation pipeline and EvalPlus~\cite{evalplus}. We report Pass@10 and Pass@1 scores for LiveCodeBench v5 and HumanEval Plus, respectively.

\paragraph{Baselines}
To better evaluate the effectiveness of our framework, we compare HEAL against the following baselines: (1) Full-shot: trained on sufficient target-domain data sampled from each domain’s training set, serving as a high-resource performance upper bound; (2) Few-shot: trained only on the few-shot target-domain samples; (3) Only-General: trained only on large-scale general-domain samples; (4) Hybrid: naively combines the target-domain samples with randomly selected general-domain samples.

\subsection{Main Results}
\label{sec:Main_res}

Table~\ref{tab:code_medicine_physics} presents the comprehensive performance of HEAL compared to various baselines across three target domains. 
% The results support a multi-dimensional analysis of the role of our framework in few-shot RLVR.
Based on the overall evaluation results, we highlight several key conclusions:

\paragraph{HEAL Significantly Improves the Performance of Few-Shot RLVR}
Our empirical results demonstrate that the HEAL framework yields substantial performance gains over the vanilla Few-shot baseline across all evaluated models and benchmarks. For instance, HEAL improves the average scores of Qwen3-1.7B-Base over the Few-shot baseline by up to 6.64\% in the Code domain. Notably, our framework consistently outperforms the Hybrid baseline, demonstrating the effectiveness of general-domain data selection and EDA.
From the perspective of entropy, HEAL effectively mitigates entropy collapse in few-shot RLVR, enabling the policy to achieve an entropy magnitude comparable to that of full-shot training, as shown in Figure~\ref{fig:preliminary_entropy_curve}.

\paragraph{HEAL Enables Few-Shot RLVR to Match or Even Surpass Full-Shot Performance}
Remarkably, using only 32 target-domain samples, HEAL matches or even surpasses the performance of models trained on 1K samples (Full-shot), with this effect particularly pronounced in the Physics and Code domains. For example, Qwen3-4B-Base with HEAL achieves average scores of 46.99\% in Physics and 53.42\% in Code, outperforming its Full-shot counterpart (44.57\% and 51.00\%, respectively). A similar trend is also evident for LLaMA3.2-3B-Instruct, which surpasses its Full-shot performance in both domains. These results further underscore HEAL’s advancement in low-resource scenarios, demonstrating its ability to achieve competitive performance with minimal target-domain data.

\paragraph{The Role of General-Domain Data}
The results of the Only-General baseline reveal that even leveraging massive general-domain samples often yields limited performance gains, consistently underperforming the Few-shot baseline (e.g., Avg. 44.69\% vs. 49.22\% in Medicine for Qwen3-4B-Base). This demonstrates that the introduced general-domain data does not directly provide domain-specific knowledge. Instead, it serves to supply more fundamental reasoning patterns. These results further highlight that the performance gains achieved by HEAL arise from its effective utilization of the diverse exploratory behaviors present in general-domain data, rather than from directly injecting domain knowledge to artificially boost performance.

\begin{figure}[!t]  
    \vspace{-4pt}
	\centering  
	\includegraphics[width=\linewidth]{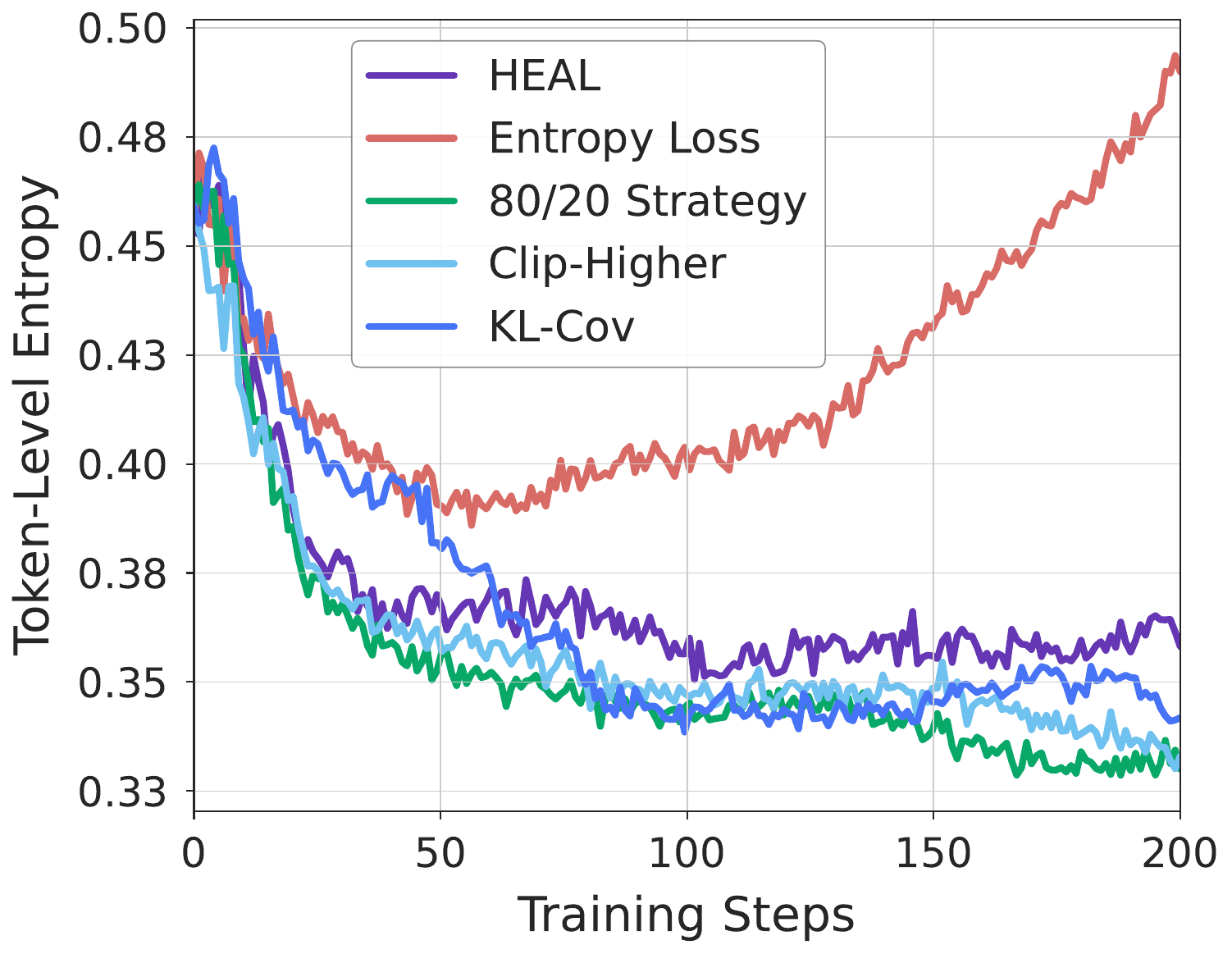}  
    % \vspace{-20pt}
	% \caption{The variation curves of average token entropy for the Qwen3-1.7B-Base model across three different target domains and under various entropy control methods.}
    \caption{Average token-level entropy during training under different entropy regularization methods and HEAL. We conduct experiments on the Qwen3-1.7B-Base model using the same training dataset.}
	\label{fig:analysis_entropy_curve}  
    % \vspace{-5pt}
\end{figure}

\begin{table}[!t]
\small
\centering
\setlength{\aboverulesep}{0pt}
\setlength{\belowrulesep}{0pt}
\setlength{\tabcolsep}{8.5pt}
\renewcommand{\arraystretch}{1.2} % 增加行高
\begin{tabular}{lccc}
\toprule
Method       & Medicine & Physics & Code \\ % 列顺序调整
\midrule
Entropy Loss  & 42.79 & 34.43 & 38.27  \\
80/20 Strategy   & 42.47 & \underline{36.36} & \underline{40.08}  \\
Clip-Higher   & 42.59 & 34.52 & 39.18  \\
KL-Cov      & \underline{44.00} & 35.58 & 38.57  \\
\addlinespace[0.5pt]
\hdashline
\addlinespace[0.5pt]
% Ours 行：背景色 + 加粗最大值
\rowcolor{TechBlue!10}
\textbf{HEAL}         & \textbf{44.67} & \textbf{37.20} & \textbf{41.28} \\
\bottomrule
\end{tabular}
% \caption{Comparison between our framework and other entropy control methods on Qwen3-1.7B-Base model under an identical training dataset configuration.}
\caption{Performance comparison between HEAL and other entropy regularization methods. We report the average score for each domain.}
\label{tab:other_entropy_method}
% \vspace{-0.2cm}
\end{table}

\subsection{Analysis}
\paragraph{Comparison with Existing Entropy Regularization Methods}

Many recent studies have proposed various entropy regularization methods to mitigate entropy collapse in RLVR. However, these methods typically do not account for data scarcity. To rigorously evaluate our framework, we compare HEAL with four representative entropy regularization baselines: \emph{Entropy Loss}~\cite{he2025skywork}, \emph{80/20 Strategy}~\cite{wang2025beyond}, \emph{Clip-Higher}~\cite{yu2025dapo}, and \emph{KL-Cov}~\cite{cui2025entropy}. Briefly, Entropy Loss directly encourages higher policy entropy through a dedicated loss term, 80/20 Strategy selects the highest-entropy tokens for policy updates, Clip-Higher increases the likelihood of low-probability exploration tokens, and KL-Cov applys KL penalty to tokens with high covariances. Details about these methods are provided in Appendix~\ref{apx:entropy_methods}.

As shown in Table~\ref{tab:other_entropy_method}, our framework consistently outperforms all baselines across the three target domains. This superior performance can be attributed to HEAL’s more effective mitigation of policy entropy collapse. As illustrated in Figure~\ref{fig:analysis_entropy_curve}, the standard Entropy Loss, which naively increases entropy without constraints, can even lead to entropy explosion in later stages of training. For the 80/20 Strategy, Clip-Higher, and KL-Cov, their ability to mitigate entropy collapse is limited in low-resource scenarios.
In contrast, HEAL leverages general-domain data as a reference to guide the policy, enabling more controlled adjustment of entropy magnitude while also utilizing learned fine-grained variation to effectively promote more diverse exploratory behaviors.

% dynamics between target and general domain,while baselines typically enforce statistical uncertainty at the token level which may introduce indiscriminate noise, our framework aligns the general-domain entropy dynamics. This ensures that the induced exploration is structured and meaningful, leading to more robust performance in data-constrained settings. More details of results on each target-domain are in Appendix~\ref{apx:more_entropy_methods}.

\definecolor{qing}{HTML}{A8D865}
\definecolor{shallowblue}{HTML}{6FC1F0}
\definecolor{deepblue}{HTML}{4402B6}

% 调整 legend 方块样式
\pgfplotsset{
    /pgfplots/ybar legend/.style={
        /pgfplots/legend image code/.code={%
           \draw[##1,/tikz/.cd,yshift=-0.25em]
            (0cm,0cm) rectangle (7pt,0.8em);},
    },
}

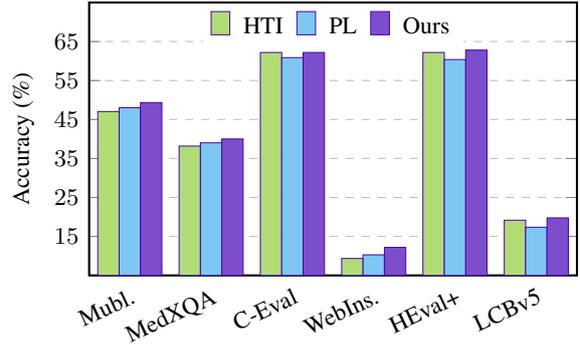
\begin{figure}[!t]
\centering
\vspace{-4pt}
\pgfplotsset{width=8cm, height=5.2cm}
\begin{tikzpicture}  

    \tikzset{every node/.style={font=\small}}
    \hspace*{-0.15cm}
    \begin{axis}[
        ybar,
        bar width=8pt, % 减小柱宽以适应三个柱子
        ymin=5, ymax=75,
        ytick={ 15, 25, 35,45,55, 65},
        major x tick style = transparent,
        enlarge x limits=0.1,
        ylabel={Accuracy (\%)},
        symbolic x coords={ Mubl., MedXQA, C-Eval, WebIns., HEval+, LCBv5},
        xtick=data,
        xticklabel style={
            rotate=25,           % 旋转45度
            anchor=north east,   % 锚点在东北方向
            inner sep=0pt,       % 减少内部间距
            yshift=0pt,         % 向下微调
            % font=\small, % 使用无衬线字体
        },
        y label style={at={(axis description cs:0.1,0.5)},anchor=south, font=\small},
        axis x line*=bottom,
        axis y line*=left,
        % 添加完整的黑色边框
        axis line style={black, thick}, % 坐标轴线条样式
        axis lines=box, % 完整的方框边界
        % 设置y轴网格线（水平虚线）
        grid=major, % 启用主要网格
        grid style={dashed, gray!50}, % 网格样式：虚线，灰色
        ymajorgrids=true, % 启用y轴主要网格线
        xmajorgrids=false, % 禁用x轴主要网格线（只保留水平线）
        legend cell align=left,
        legend style={
            at={(0.5,0.83)},
            anchor=south,
            legend columns=3,
            draw=none,
            font=\small,
            /tikz/column sep=3pt,
        },
    ]
        % HI - 使用cinnamon颜色，向右偏移-6pt
        \addplot[ybar, fill=qing!90!white, draw=deepblue, line width=0.1pt, 
                bar shift=-8pt] coordinates {
            (Mubl., 47.04) (MedXQA, 38.18) (C-Eval, 62.19) (WebIns., 9.35) (HEval+, 62.19) (LCBv5, 19.17) 
        };
        % PL - 使用bananayellow颜色，不偏移
        \addplot[ybar, fill=shallowblue!90!white, draw=deepblue, line width=0.1pt,
                bar shift=0pt] coordinates {
             (Mubl., 48.02) (MedXQA, 39.01) (C-Eval, 60.87) (WebIns., 10.28) (HEval+, 60.38) (LCBv5, 17.36)
        };
        % Ours - 使用rondom颜色，向左偏移6pt
        \addplot[ybar, fill=deepblue!70!white, draw=deepblue, line width=0.2pt,
                bar shift=8pt] coordinates {
             (Mubl., 49.34) (MedXQA, 40) (C-Eval, 62.19) (WebIns., 12.20) (HEval+, 62.8) (LCBv5, 19.76)
        };
        
        \legend{
            HTI,
            PL,
            Ours,
        }
    \end{axis}
\end{tikzpicture}
\caption{Performance comparison using different similarity functions. Experiments are conducted on the Qwen3-1.7B-Base model.}
\label{fig:distance}
\end{figure}

\paragraph{Impact of Different Similarity Measures for Entropy Dynamics}

In addition to the KL divergence used in our final implementation, we also experiment with two alternative strategies: High-entropy Tokens Intersection (HTI) and Pearson Linear (PL) similarity. Briefly, HTI measures the overlap between high-entropy tokens, while PL computes the correlation of slopes after linearly fitting the trajectory-level entropy dynamics. Implementation details are provided in Appendix~\ref{appendix:distance_func}.
As shown in Figure~\ref{fig:distance}, the KL divergence used in our framework consistently outperforms the other two methods, demonstrating its superior ability to capture fine-grained differences in trajectory-level entropy dynamics. This precise measurement enables more effective alignment, allowing the policy to acquire more diverse exploratory behaviors from the general domain.

\paragraph{Effect of Training Data Scale}
\label{subsec:scaling}
Figure~\ref{fig:Scaling} illustrates the impact of training data scale in both the target and general domains on the performance of the Qwen3-1.7B-Base model. 
In the target-domain experiment, HEAL surpasses the Hybrid baseline at every data size and approaches or even exceeds the Full-shot upper bound with as few as 32 samples, highlighting its ability to effectively leverage scarce samples. In the general-domain experiment, the improvement of HEAL is also substantially stronger than that of the Hybrid baseline, indicating that it can more effectively transfer fundamental reasoning patterns and diverse exploratory behaviors from general-domain data to the target domain.

\definecolor{blue}{HTML}{4402B6}
\usetikzlibrary{pgfplots.groupplots}
\begin{figure}[t]
    \vspace{-6pt}
    \centering
    \begin{tikzpicture}
        \begin{groupplot}[
            group style={
                group size=2 by 1, % 一行两列
                horizontal sep=0.6cm, % 子图之间的水平间距
                vertical sep=1.25cm, % 子图之间的垂直间距
                group name=myplots, % 为组命名，便于后续引用
            },
            width=0.275\textwidth, % 每个子图的宽度
            height=0.3\textwidth, % 每个子图的高度
            grid=both, % 显示网格
            major grid style={line width=0.4pt, draw=gray!50},
            minor grid style={line width=0.25pt, draw=gray!20},
            tick label style={font=\small},
            label style={at={(axis description cs:0.25,0.5)},anchor=south, font=\small},
            legend cell align=left,
            title style={yshift=-1ex, font=\small},
            % 统一的轴设置
            axis lines=box, % 完整的方框边界
            axis line style={black, thick}, % 坐标轴线条样式
        ]
        % 第一个子图
        \nextgroupplot[
            ylabel={Accuracy (\%)}, % y轴标签
            xmin=0, xmax=3, % 设置x轴范围，包含所有数据点
            ymin=34, ymax=42,
            xtick={0,1,2,3}, % 显示7个刻度（0到7）
            xticklabels={0,2,8,32,128}, % 自定义标签
            % 调整y轴标签位置，使其紧贴刻度
            % ylabel style={at={(ticklabel* cs:0.5)}, xshift=-0.8cm, anchor=center, rotate=90},
            % legend pos=north west, % 暂时保留图例位置，稍后我们将移动它
            legend style={
                at={(1.05,-0.35)},
                anchor=south,
                legend columns=3,
                draw=none,
                font=\small,
                /tikz/column sep=3pt,
            },
            title={(a) Target Domain} % 子图标题
        ]
            \addplot[
                color=olive,
                line width=1pt
            ] coordinates {
                (0,  40.05)
                (1,  40.05)
                (2,    40.05)
                (3,  40.05)
            };
            \addlegendentry{Full-shot}
            \addplot[
                color=cyan,
                mark=o,
                line width=1pt
            ] coordinates {
                (0,   34.45)
                (1, 35.10)
                (2,   35.77)
                (3,   37.27)
            };
            \addlegendentry{Hybrid}
            \addplot[
                color=blue,
                mark=o,
                line width=1pt
            ] coordinates {
                (0,  34.45)
                (1,  35.75)
                (2,  37.18)
                (3,  41.05)
            };
            \addlegendentry{HEAL}
        
        % 第二个子图
        \nextgroupplot[
            xmin=0, xmax=3, % 设置x轴范围，包含所有数据点
            ymin=34, ymax=42,
            xtick={0,1,2,3}, % 显示7个刻度（0到7）
            xticklabels={0,96,192,384,768}, % 自定义标签
            % ytick pos=right, % 将y轴刻度放在右侧
            % yticklabel style={
            %     font=\small,
            %     xshift=0pt, % 稍微向右偏移，避免与网格线重叠
            % },
            yticklabels={},
            axis lines=box, % 完整的方框边界
            title={(b) General Domain} % 子图标题
        ]
            \addplot[
                color=cyan,
                mark=o,
                line width=1pt
            ] coordinates {
                (0,  36.21)
                (1,  36.37)
                (2,  36.91)
                (3,  37.27)
            };
            % 不添加图例，使用统一的图例
            \addplot[
                color=blue,
                mark=o,
                line width=1pt
            ] coordinates {
                (0,  36.21)
                (1,  37.61)
                (2,  38.85)
                (3,  41.05)
            };
            \addplot[
                color=olive,
                line width=1pt
            ] coordinates {
                (0,  40.05)
                (1,  40.05)
                (2,  40.05)
                (3,  40.05)
            };
        \end{groupplot}
    \end{tikzpicture}
    \caption{Average accuracy of the Qwen3-1.7B-Base model across three target domains with respect to different data sizes: (a) varying the target-domain data size while keeping the general-domain data fixed at 384 samples; (b) varying the general-domain data size while keeping the target-domain data fixed at 32 samples.
    }
    \label{fig:Scaling}
    % \vspace{-10pt}
\end{figure}
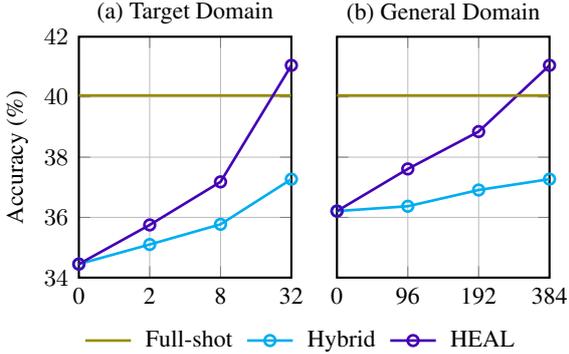
\paragraph{Adaptation to Math Domain}
Although HEAL is specifically designed for low-resource domains, it inherently offers a data-efficient solution under limited computational resources. Therefore, we extend our framework to a high-resource domain (i.e., Math) to assess the efficiency of HEAL. As shown in Table~\ref{tab:math}, HEAL still demonstrates consistently strong performance. Notably, compared with Full-shot training, HEAL matches or even surpasses its performance while using only 40\% of the training data. For instance, on AMC23 and Minerva, HEAL exceeds the Full-shot baseline by 3.49\% and 2.57\%, respectively. These results underscore HEAL’s versatility as a data-efficient RLVR approach for high-resource domains.

\begin{table}[!t]
\small
\setlength{\tabcolsep}{3.5pt} % 列数较少，可以稍微放宽间距增加可读性
% \centering
\setlength{\aboverulesep}{0pt}
\setlength{\belowrulesep}{0pt}
\renewcommand{\arraystretch}{1.15} % 增加行高，美观且适配背景色
\begin{tabular}{lcccc} % 修正列数定义为 5 列
\toprule
 & AMC23 & Math500 & Minerva & Olym. \\
\midrule
Qwen3-1.7B-Base                   & 30.94 & 55.6 & 13.24 & 21.48 \\
\textcolor{gray}{$\hookrightarrow$}~~Full-shot        & \underline{35.31} & \textbf{66.2} & \underline{19.12} & \textbf{29.48}\\
\textcolor{gray}{$\hookrightarrow$}~~Few-shot          & 34.06 & 58.4 & 17.28 & 23.70\\
\textcolor{gray}{$\hookrightarrow$}~~Only-General & 29.38 & 62.8 & 16.91 & \underline{26.07} \\
\textcolor{gray}{$\hookrightarrow$}~~Hybrid & 32.50 & 60.2 & 17.65 & 24.15 \\
\addlinespace[0.5pt]
\hdashline
\addlinespace[0.5pt]
\rowcolor{TechBlue!10} % 科技蓝背景
\textcolor{gray}{$\hookrightarrow$}~~HEAL  & \textbf{38.80} & \underline{64.4} & \textbf{21.69} & 25.78  \\
\bottomrule
\end{tabular}
\caption{Performance of Qwen3-1.7B-Base across four Math domain benchmarks. Details of datasets are in Appendix~\ref{apx:more_datasets}, and additional results are in Appendix~\ref{apx:more_math}.}
\label{tab:math}
\end{table}

\begin{table}[!t]
\centering
\small
% --- 核心：局部消除booktabs产生的断点并调整行高 ---
\setlength{\tabcolsep}{5.5pt} 
\setlength{\aboverulesep}{0pt} 
\setlength{\belowrulesep}{0pt} 
\renewcommand{\arraystretch}{1.1} % 增加行高以弥补消除的sep，使表格不拥挤

\label{tab:ablation}
\begin{tabular}{c @{\hspace{4pt}} c @{\hspace{5pt}} c !{\vrule width 0.5pt} ccc} % !{\vrule...} 定义细竖线
\toprule
\multicolumn{2}{c}{Data Selection} & \multirow{2}{*}{EDA} & \multirow{2}{*}{Med.} & \multirow{2}{*}{Phy.}  & \multirow{2}{*}{Code} \\ 
% 这里的 r{0.4em} 留出一点空隙给竖线
\cmidrule(r{0.4em}){1-2} 
$\mathrm{Uncertainty}$ & $\mathrm{Diversity}$ & & & & \\ 
\midrule
$\times$     & $\times$     & $\times$      & 42.02 & 33.94 & 35.85 \\ 
$\checkmark$ & $\times$     & $\times$      & 41.49 & 34.28 & 36.45 \\ 
$\times$     & $\checkmark$ & $\times$      & 42.16 & 34.53 & 37.04 \\ 
$\times$     & $\times$     & $\checkmark$  & \underline{43.80} & \underline{36.44} & \underline{39.47} \\ 
$\checkmark$ & $\checkmark$ & $\times$      & 43.04 & 35.34 & 37.36 \\ 
\addlinespace[0.5pt]
\hdashline
\addlinespace[0.5pt]
\rowcolor{TechBlue!10}
$\checkmark$ & $\checkmark$ & $\checkmark$   & \textbf{44.67} & \textbf{37.20} & \textbf{41.28} \\ 
\bottomrule
\end{tabular}
\caption{Ablation study of the Qwen3-1.7B-Base model across the Medicine (Med.), Physics (Phy.), and Code domains. Average scores are reported.}
% Data Selection: Adopts our general-domain data selection strategy with $\mathrm{Difficulty}$ and $\mathrm{Diversity}$ criterion. EDA: Incorporates our entropy dynamics alignment reward into RLVR training.}
\label{tab:ablation}
\end{table}

\subsection{Ablation Study}
The ablation study of our framework is summarized in Table~\ref{tab:ablation}. We systematically evaluate the contributions of the proposed general-domain data selection strategy and Entropy Dynamics Alignment (EDA).
When none of the components is applied, the model exhibits limited performance across all three target domains. Introducing either uncertainty-based or diversity-based data selection leads to only marginal improvements, while combining the two yields more consistent gains, suggesting that these criteria are complementary in identifying high-value general-domain samples. Notably, enabling EDA alone results in a substantial performance boost, underscoring its critical role in mitigating entropy collapse in few-shot RLVR.
Overall, the results demonstrate that both high-quality general-domain data selection and EDA are essential to the effectiveness of HEAL, and that their combination produces clear synergistic gains under low-resource scenarios.

% \subsection{Adaptation to Math Domain}

\section{Conclusion}
In this work, we proposed HEAL, a novel framework for mitigating entropy collapse in few-shot RLVR. By integrating selected general-domain data and introducing Entropy Dynamics Alignment, HEAL effectively encourages more diverse policy behaviors. Extensive experiments across multiple domains demonstrate that HEAL consistently improves few-shot RLVR performance, even matching or surpassing full-shot training with significantly fewer samples. These results highlight HEAL’s effectiveness as a data-efficient approach for RLVR in low-resource scenarios.

\section*{Limitations}
\addcontentsline{toc}{section}{Limitations}
Despite our results are promising, several limitations remain. First, due to computational constraints, HEAL has not been evaluated on larger models, leaving its scalability to be validated. Second, incorporating general-domain data introduces some overhead, but it is limited since only a small amount is used, and we consider it acceptable, as our framework can match or even surpass performance using much more target-domain data. 
%Third, the EDA component in our framework involves inter- and intra-domain similarity computations, which adds additional cost. A promising solution is to compute the EDA reward in parallel with the importance sampling phase, which we leave for future work.

\section*{Acknowledgements}
\addcontentsline{toc}{section}{Acknowledgements}
The project was supported by National Key R\&D Program of China (No. 2022ZD0160501), Natural Science Foundation of Fujian Province of China (No. 2024J011001), and the Public Technology Service Platform Project of Xiamen (No.3502Z20231043). We also thank the reviewers for their insightful comments.

% Bibliography entries for the entire Anthology, followed by custom entries
%\bibliography{anthology,custom}
% Custom bibliography entries only
\bibliography{custom}

\appendix
% \newpage
\section{Implementation Details}
\label{sec:appendix A}

\subsection{Token-Level Entropy Calculation}
\label{token-entropy}
Token-level entropy quantifies the uncertainty of the model's token-generation distribution at a given decoding step. Formally, at decoding step $t$, the entropy is defined as: 
\begin{equation}
\label{eq:token-entorpy}
\mathcal{H}_t = -\sum_{j=1}^{\left|V \right|} p_{t,j} \log p_{t,j},
\end{equation}
where \( p_t = \pi_\theta(\cdot \mid q, o_{<t}) = \emph{Softmax}(\frac{z_t}{T}) \) is the probability distribution over the vocabulary $V$. Here, \( \pi_\theta \) is the model parameterized by \( \theta \), \( q \) is the input query, \( o_{<t} \) denotes previously generated tokens, \( z_t \) represents the pre-softmax logits, and \( T \) is the temperature.
In this work, the term \emph{token-level entropy} refers specifically to the entropy of the entire vocabulary distribution at a given decoding step, and not to any individual token instance. 

Following~\citet{cui2025entropy}, we adopt policy entropy to quantify the predictability or randomness inherent in the actions selected by an agent. Specifically, given policy model $\pi_{\theta}$, training dataset $\mathcal{D}$, we measure the average token-level entropy of the policy model on training data, which is defined as follows:
\begin{equation}
\small
\begin{aligned}
\mathcal{H}(\pi_{\theta}, \mathcal{D}) 
= - \frac{1}{|\mathcal{D}|} \sum_{x \in \mathcal{D}} \frac{1}{|o|} \sum_{t=1}^{|o|}\mathbb{E}_{o_t \sim \pi_{\theta}}\left[\log \pi_{\theta}(o_t|o_{<t}, q) \right]
\end{aligned}
\end{equation}
Such entropy quantifies the uncertainty level of the policy on current prompts and is widely known as sequence-level mean entropy. In practice, we calculate the entropy for each batch of prompts randomly sampled from the training dataset.

\subsection{Details of the Entropy Dynamics Similarity Functions}
\label{appendix:distance_func}
\paragraph{Proposed Similarity Measure via KL Divergence}
To facilitate the comparison of entropy dynamics with varying lengths, we employ an interpolation-based alignment method.
Specifically, given two entropy dynamics $\tau_i$ and $\tau_j$ (assuming $\left| \tau_i \right| < \left| \tau_j \right|$), we apply \emph{nearest-neighbor interpolation} to $\tau_i$ to align its length with $\tau_j$, then we get the \(\tau_i'\) with length $\left| \tau_j \right|$. Subsequently, to prioritize the fluctuation patterns of token entropy over their absolute magnitudes, we normalize the entropy dynamics using the \emph{Softmax} operation:
\begin{equation}
    \hat{\tau}_i = \textit{Softmax}(\tau_i'), \quad  \hat{\tau}_j = \textit{Softmax}(\tau_j ).
\end{equation}
Finally, we define the similarity between the two entropy dynamics as:
\begin{equation}
\small
\label{eq:ourds}
    s_{\text{Ours}}(\hat{\tau}_i, \hat{\tau}_j) = -\mathbb{D}_{\text{KL}}(\hat{\tau}_i || \hat{\tau}_j) = -\sum_{t=1}^{\left| \tau_j \right|} \hat{\tau}_{i,t} \log \frac{\hat{\tau}_{i,t}}{\hat{\tau}_{j,t}},
\end{equation}
where $\mathbb{D}_{\text{KL}}(\cdot,\cdot)$ denotes the KL divergence. Since KL divergence is a non-negative measure of difference where a larger value implies lower similarity, we negate it to ensure that a higher $s_{\text{Ours}}$ indicates greater similarity.

In addition to the above implementation adopted in our framework, we explored alternative metrics to verify that the aforementioned similarity measure is relatively optimal. Specifically, we experimented with two other methods for comparison, as described below:
\paragraph{High-entropy Tokens Intersection Similarity}
%distance draws inspiration from the histogram intersection method, a conventional technique for measuring the similarity between two distributions. Specifically, HTI 
The High-entropy Tokens Intersection (HTI) is specialized for entropy dynamics through an analysis of overlap among the key segments that govern exploratory behavior. After aligning two entropy dynamics $\tau_i$ and $\tau_j$ to a common length $N$ via interpolation, we identify the top 20\% of tokens with the highest entropy. Let $\mathcal{I}_i^{top}$ and $\mathcal{I}_j^{top}$ be the sets of indices corresponding to these high-entropy tokens in trajectories. The similarity is computed as:
\begin{equation}
\small
    s_{\text{HTI}}(\tau_i, \tau_j) = \sum_{t=1}^N \min \left( \tau_{i,t} \cdot \mathbb{I}_{t \in \mathcal{I}_i^{top}}, \tau_{j,t} \cdot \mathbb{I}_{t \in \mathcal{I}_j^{top}}\right),
\end{equation}
where $\mathbb{I}_{(\cdot)}$ denotes the indicator function. This approach provides an efficient measure of coarse-grained alignment in entropy dynamics, primarily highlighting shared high-uncertainty behaviors. However, it may not capture fine-grained structural details in the lower-entropy segments.

\paragraph{Pearson Linear Similarity}
The Pearson Linear (PL) similarity characterizes the global linear trend of an entropy dynamic by modeling the relationship between token entropy and its sequential index using linear regression. For each entropy dynamic $\tau$, we extract the slope $k_\tau$ of the fitted line and the Pearson correlation coefficient $\delta_\tau$. The similarity between two entropy dynamics $\tau_i$ and $\tau_j$ is defined by integrating their angular proximity and linear consistency:
\begin{equation}
\small
    s_{\text{PL}}(\tau_i, \tau_j)  = \left| \cos(\arctan k_{\tau_i} - \arctan k_{\tau_j}) \cdot \delta_{\tau_i} \cdot \delta_{\tau_j} \right|,
\end{equation}
where $\cos(\arctan k_{\tau_i} - \arctan k_{\tau_j})$ quantifies the angular difference between the two fitted lines, while the product $\delta_{\tau_i} \cdot \delta_{\tau_j}$ accounts for the reliability of the linear trends. A significant advantage of the PL similarity is that it does not require length-alignment operations, such as interpolation, maintaining low computational complexity. However, its reliance on linear approximations may limit its ability to capture highly non-linear or local fluctuations within the entropy dynamics.

\subsection{Details of Entropy Regularization Baselines}
\label{apx:entropy_methods}

\paragraph{Entropy Loss~\cite{he2025skywork}} This method directly incorporates the principles of maximumm entropy RL by augmenting the GRPO objective with an entropy regularization term. Formally:
\begin{equation}
\mathcal{L}_{\text{AEC}}(\theta)  = -\frac{\alpha}{\left|y_i\right|G}\sum_{i=1}^{G} \sum_{t=1}^{\left|y_i\right|}\mathcal{H}_t^i,
\end{equation}
where $\alpha$ is a hyperparameter controlling the strength of the entropy regularization, and $\mathcal{H}_t^i$ denotes the token-level entropy at step $t$ for response 
$y_i$, as defined in Equation~\ref{eq:token-entorpy}.
Then the final loss function of Entropy baseline is defined as:
\begin{equation}
\mathcal{L}_{\text{Entropy loss}}(\theta)  = \mathcal{L}_{\text{GRPO}}(\theta) + \mathcal{L}_{\text{AEC}}(\theta).
\end{equation}
This formulation ensures a lower bound on policy entropy, thereby preserving exploration capability, mitigating entropy collapse, and maintaining learning plasticity during training.

\paragraph{80/20 Strategy~\cite{wang2025beyond}}

Standard GRPO treats all tokens uniformly, often allowing low-entropy tokens to overshadow critical updates. To address this, 80/20 Strategy employs a mask $M_{i,t} = \mathbb{I}(\mathcal{H}_t^i \geq \delta_\gamma)$ to restrict updates to the top $\gamma$ (e.g., 20\%) high-entropy tokens. The objective is formulated as:
\begin{equation}
\small
\begin{aligned}
    \mathcal{L}_{\text{80/20}}(\theta) = -\frac{1}{N_{HE}} \sum_{i=1}^{G} \sum_{t=1}^{\left|y_i\right|} M_{i,t}  \min \left( \rho_{i,t} \hat{A}_i, \rho_{i,t}^{\text{clip}} \hat{A}_i \right) \\
     + \beta \mathbb{D}_{\text{KL}}(\pi_\theta || \pi_{\text{ref}}),
\end{aligned}
\end{equation}
where $N_{\text{HE}} = \sum_{i,t} M_{i,t}$ denotes the total number of high-entropy tokens in the batch, and $\rho_{i,t}$ is the token-level probability ratio. By prioritizing ``forking'' tokens that represent pivotal logical transitions, this method prevents the KL regularization from being dominated by trivial linguistic patterns, thereby accelerating convergence and enhancing reasoning robustness.

\paragraph{Clip-Higher~\cite{yu2025dapo}}
As a core component of DAPO, it encourages higher entropy by using an asymmetric clipping range in the PPO objective. It addresses the observation that standard symmetric clipping disproportionately restricts probability increases for unlikely tokens. Specifically, the clipping range is decoupled into two parameters: a smaller $\epsilon_{\text{low}}$ to prevent collapse, and a larger  $\epsilon_{\text{high}}$ to allow greater flexibility for “\emph{exploration tokens}”. 
The clipped ratio in GRPO objective is modified as: 
\begin{equation}
    \rho_i^{\text{clip-higher}} = \text{clip}(\rho_i, 1-\epsilon_{\text{low}}, 1+\epsilon_{\text{high}}).
\end{equation}
This strategy mitigates entropy collapse and promotes the generation of more diverse samples by enhancing the policy's exploration capabilities.

\paragraph{KL-Cov~\cite{cui2025entropy}}
This method counteracts entropy collapse by applying a selective KL‑divergence penalty to tokens exhibiting high covariance between their log‑probabilities and advantages. Specifically, the strategy selects token indices $ \mathcal{I}_{\text{KL}}$ corresponding to the top-$k$ covariance values:
\begin{equation}
\small
    \mathcal{I}_{\text{KL}} = \{t \mid \text{Rank}(\text{Cov}(o_t)) \leq k \cdot N_T\},
\end{equation}
where $k \ll 1$ denotes the proportion of tokens targeted for regularization, and $N_T$ is the total number of tokens. For tokens within this set, a KL penalty is imposed to regularize the divergence between the current policy $\pi_\theta$ and the rollout policy $\pi_{\theta_{\text{old}}}$. The resulting policy loss is formulated as:
\begin{equation}
\small
L_{\text{KL-Cov}}(\theta) = 
\begin{cases} 
\mathbb{E}_t [ \rho_t A_t ], & t \notin  \mathcal{I}_{\text{KL}} \\
\mathbb{E}_t [ \rho_t A_t - \beta \mathbb{D}_{\text{KL}}(\pi_{\theta_{\text{old}}}\| \pi_\theta) ], & t \in \mathcal{I}_{\text{KL}}
\end{cases}
\end{equation}
where the importance sampling ratio $\rho_t$, $A_t$ is the advantage estimate, and $\beta$ is a hyperparameter controlling the weight of the KL penalty. By selectively penalizing tokens that exhibit high covariance, KL-Cov constrains the policy update within a stable trust region, thereby preserving exploration diversity and preventing premature convergence.

\begin{table}[!t]
\small
\centering
\renewcommand{\arraystretch}{1.125} 
\setlength{\aboverulesep}{0pt}
\setlength{\belowrulesep}{0pt}
\setlength{\tabcolsep}{4pt}
\renewcommand{\arraystretch}{1.3} % 增加行高
\label{tab:hyperparams}
\begin{tabular}{lccc}
\toprule
Method & Key Hyperparameter & Symbol & Value \\
\midrule
Entropy Loss & Entropy coefficient & $\alpha$ & 0.001 \\
\hline
80/20 Strategy & \thead{The fraction of \\high-entropy tokens} & Top-$\gamma$ & 20\% \\
\hline
\multirow{2}{*}{Clip-higher} & Upper threshold & $\varepsilon_{\text{high}}$ & 0.28 \\
& Lower threshold & $\varepsilon_{\text{low}}$ & 0.2 \\
\hline
\multirow{2}{*}{KL-Cov} & \thead{The fraction of \\ high-covariance} & Top-$k$ & 0.02\% \\
& KL penalty coefficient & $\beta$ & 1 \\
\bottomrule
\end{tabular}
\caption{To ensure a solid comparison, we implemented these entropy-based methods using the default hyperparameter configurations specified in their original works, as they are inherently agnostic to dataset scale.}
\end{table}

\begin{figure*}[!t]    
    % \vspace{-5pt}
	\centering  
	\includegraphics[width=\linewidth]{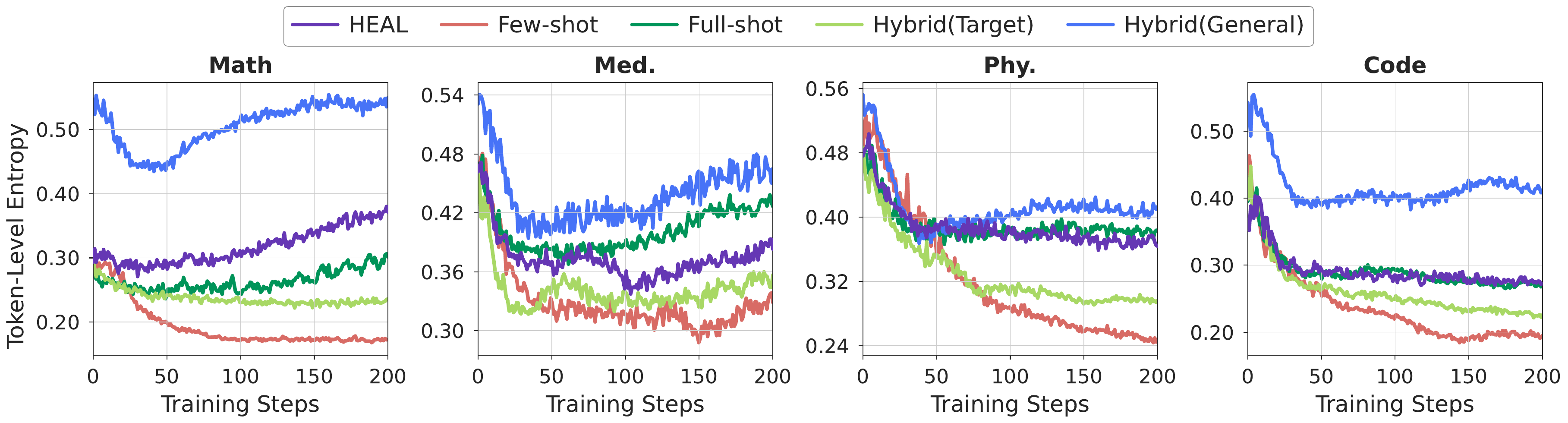}  
    % \vspace{-20pt}
	% \caption{The variation curves of average token entropy for the Qwen3-1.7B-Base model across three different target domains and under various entropy control methods.}
    \caption{Details of separate average token-level entropy curves on Math,  Medicine (Med.), Physics (Phy.), and Code domains. Evaluated on the Qwen3-1.7B-Base model under different datasets settings baselines and our framework HEAL. Hybrid(Target) and Hybrid(General) indicate few-shot RLVR augmented with randomly sampled general-domain data, evaluated on target- and general-domain data.}
	\label{fig:more_details_en_main}  
    % \vspace{-5pt}
\end{figure*}

\section{Details of Training and Evaluation Datasets}
\label{sec:appendix B}
\paragraph{Details of Medicine Datasets}
MedBullets~\cite{chen-etal-2025-benchmarking} comprises high-quality multiple-choice questions, providing a rigorous foundation for clinical knowledge. MedXpertQA~\cite{zuo2025medxpertqa} features expert-level medical queries sourced from professional exams and clinical cases, designed to challenge the model's complex reasoning and domain-specific problem-solving skills. To evaluate expert-level clinical reasoning, we integrate MedBullets and MedXpertQA.

\paragraph{Details of Physics Datasets}
WebInstruct~\cite{ma2025generalreasoner} is a large-scale, high-quality instruction-following dataset curated through extensive web mining from Common Crawl. And C-Eval~\cite{huang2023ceval} is a comprehensive evaluation suite spanning multiple subjects. We specifically employ their physics-related subsets to assess the model's reasoning capabilities across diverse physics problems of varying difficulty levels.

\paragraph{Details of Code Datasets}
LiveCodeBench~\cite{jain2024livecodebench} collects problems from competitive programming platforms with a focus on contamination-free evaluation. We utilize data from versions v1--v4 to construct our training set, while employing LiveCodeBench v5 as an evaluation benchmark. HumanEval Plus (HEval+)~\cite{evalplus} serves as a rigorous expansion of the original HumanEval dataset~\cite{chen2021codex}, featuring augmented test cases.  We employ them to assess coding capabilities.

\paragraph{Details of Math Datasets}
\label{apx:more_datasets}
 Following ~\citet{wang2025reinforcement}, we select 1K examples from DeepScaleR~\cite{luo2025deepscaler} as the full-shot dataset for the math domain. For evaluation, we adopt four benchmarks: AMC23~\cite{amc23} comprises 40 challenging problems sourced from prestigious secondary school mathematics competitions. Math500~\cite{lightman2023let} provides a computationally efficient evaluation through a curated subset of the MATH benchmark's test partition~\cite{verma2025measuring}. The math portion of Olympiad~\cite{he2024olympiadbench} contains international competition problems that demand expert-level logical deduction. Minerva~\cite{lewkowycz2022solving} features undergraduate STEM problems from MIT OpenCourseWare designed to evaluate complex multi-step reasoning. We report results using the Avg@4 metric, except for the smaller AMC23 set, which adopts Avg@8.

\definecolor{TechBlue}{RGB}{0, 123, 255}
\begin{table*}[!t]
\small
% 调整行高，让表格不拥挤（替代 addlinespace）
\renewcommand{\arraystretch}{1.1} 
% 关键设置：消除 booktabs 对竖线的截断，使竖线贯穿全表
\setlength{\aboverulesep}{0pt}
\setlength{\belowrulesep}{0pt}
\setlength{\tabcolsep}{10pt}
\centering

% 列定义：增加了Data Size列
\begin{tabular}{l | c @{\hspace{4pt}} c | c c c c | c}
\toprule
& \multicolumn{2}{c|}{Data Size} & \multicolumn{4}{c|}{Math} & \\
\cmidrule(lr){2-3} \cmidrule(lr){4-7}
 & $\left| \mathcal{D}_{tgt} \right|$ & $\left|\mathcal{D}_{gen}\right|$ & AMC23 & Math500 & Minerva & Olympiad & Avg. \\
\midrule
Qwen3-1.7B-Base                                  &          &           & 30.94 & 55.6 & 13.24 & 21.48 & 30.32 \\
\textcolor{gray}{$\hookrightarrow$} Full-shot      & 1K & 0        & \underline{35.31} & \textbf{66.2} & \underline{19.12} & \textbf{29.48} & \underline{37.53} \\
\textcolor{gray}{$\hookrightarrow$} Few-shot         & 32       & 0        & 34.06 & 58.4 & 17.28 & 23.70 & 33.36 \\
\textcolor{gray}{$\hookrightarrow$} Only-General    & 0        & 10K & 29.38 & 62.8 & 16.91 & \underline{26.07} & 33.79 \\
\textcolor{gray}{$\hookrightarrow$} Hybrid   & 32       & 384      & 32.50 & 60.2 & 17.65 & 24.15 & 33.63 \\
% 科技蓝背景，!10 代表极高透明度(仅10%颜色浓度)
\rowcolor{TechBlue!10}
\textcolor{gray}{$\hookrightarrow$} \textbf{HEAL}                    & 32       & 384 & \textbf{38.80} & \underline{64.4} & \textbf{21.69} & 25.78 & \textbf{37.67} \\
\hdashline
Qwen3-4B-Base                                    &          &           & 42.19 & 66.2 & 20.22 & 34.67 & 40.82 \\
\textcolor{gray}{$\hookrightarrow$} Full-shot       & 1K & 0        & \textbf{54.37} & \underline{77.4} & \underline{33.82} & \underline{38.81} & \underline{51.10} \\
\textcolor{gray}{$\hookrightarrow$} Few-shot     & 32       & 0        & 50.00 & 73.4 & 31.98 & 37.48 & 48.22 \\
\textcolor{gray}{$\hookrightarrow$} Only-General   & 0        & 10K & 47.81 & 69.2 & 24.26 & 34.37 & 43.91 \\
\textcolor{gray}{$\hookrightarrow$} Hybrid   & 32       & 384      & 51.88 & 74.4 & 32.72 & 38.20 & 49.30 \\
% 科技蓝背景
\rowcolor{TechBlue!10}
\textcolor{gray}{$\hookrightarrow$} \textbf{HEAL}                    & 32       & 384 & \underline{52.81} & \textbf{77.6} & \textbf{35.29} & \textbf{39.11} & \textbf{51.52} \\
\hdashline
Qwen3-8B-Base                                    &        &         & 45.63 & 63.00 & 18.01 & 33.33 & 39.99 \\
\textcolor{gray}{$\hookrightarrow$} Full-shot & 1K       & 0         & \textbf{63.75} & \underline{78.40} & \underline{35.66} & \textbf{42.96} & \textbf{55.19} \\
\textcolor{gray}{$\hookrightarrow$} Few-shot  & 32       & 0         & 58.13 & 77.20 & 34.56 & 41.04 & 52.73 \\
\textcolor{gray}{$\hookrightarrow$} Only-General     & 0        & 10K      & 53.10 & 68.60 & 23.90 & 37.63 & 45.81 \\
\textcolor{gray}{$\hookrightarrow$}    Hybrid   & 32       & 384       & 54.06 & 73.80 & 29.04 & 39.11 & 49.00 \\
\rowcolor{TechBlue!10}
\textcolor{gray}{$\hookrightarrow$} \textbf{HEAL}            & 32       & 384       & \underline{59.69} & \textbf{78.60} & \textbf{38.60} & \underline{41.48} & \underline{54.59} \\
\hdashline
LLaMA3.2-3B-Instruct                             &          &           & 25.00 & 40.8 & 15.81 & 13.19 & 23.70 \\
\textcolor{gray}{$\hookrightarrow$} Full-shot      & 1K & 0        & \textbf{30.42} & \underline{49.4} & \textbf{22.06} & 17.04 & \textbf{29.73} \\
\textcolor{gray}{$\hookrightarrow$} Few-shot    & 32       & 0        & 26.88 & 46.0 & \underline{19.49} & \textbf{17.48} & 27.46 \\
\textcolor{gray}{$\hookrightarrow$} Only-General     & 0        & 10K & 25.31 & 47.6 & 16.91 & 17.33 & 26.79 \\
\textcolor{gray}{$\hookrightarrow$} Hybrid     & 32       & 384      & 26.25 & 46.8 & 18.01 & 16.59 & 26.91 \\
% 科技蓝背景
\rowcolor{TechBlue!10}
\textcolor{gray}{$\hookrightarrow$} \textbf{HEAL}                    & 32       & 384 & \underline{28.44} & \textbf{49.8} & 18.75 & \underline{17.19} & \underline{28.55} \\
\bottomrule
\end{tabular}
\caption{Performance comparison on Math benchmarks. Average scores are computed across the four math datasets. Best results are in bold and the second best are underlined.}
\label{tab:math_only}
% \vspace{-0.25cm}
\end{table*}

\definecolor{TechBlue}{RGB}{0, 123, 255}
\begin{table*}[!t]
\small
% 调整行高，让表格不拥挤
\renewcommand{\arraystretch}{1.125} 
% 关键设置：消除 booktabs 对竖线的截断
\setlength{\aboverulesep}{0pt}
\setlength{\belowrulesep}{0pt}
\setlength{\tabcolsep}{4.15pt}
\centering

% 列定义：与参考表格一致的列结构
\begin{tabular}{l | c c | c c c | c c c | c c c}
\toprule
& \multicolumn{2}{c|}{Data Size} & \multicolumn{3}{c|}{Medicine} & \multicolumn{3}{c|}{Physics} & \multicolumn{3}{c}{Code} \\
\cmidrule(lr){2-3} \cmidrule(lr){4-6} \cmidrule(lr){7-9} \cmidrule(lr){10-12}
 & $\left| \mathcal{D}_{tgt} \right|$ & $\left|\mathcal{D}_{gen}\right|$ & MBul. & MedXQA & Avg. & C-Eval & WebIns. & Avg. & HEval+ & LCBv5 & Avg. \\
\midrule
Entropy Loss                       & 32  &   384 & 47.70 & 37.88 & 42.79 & 59.92 & 8.94 & 34.43 & 60.98 & 15.56 & 38.27 \\
80/20 Strategy                              & 32  &   384 & 47.37 & 37.57 & 42.47 & \textbf{62.95} & \underline{9.76} & \underline{36.36} & \underline{62.19} & \underline{17.96} & \underline{40.08} \\
Clip-Higher                        & 32  &   384 & 46.38 & 38.79 & 42.59 & 60.49 & 8.54 & 34.52 & 61.59 & 16.76 & 39.18 \\
KL-Cov                             & 32  &   384 & \underline{48.68} & \underline{39.31} & \underline{44.00} & 61.81 & 9.35 & 35.58 & 60.98 & 16.16 & 38.57 \\
\addlinespace[0.5pt]
\hdashline
\addlinespace[0.5pt]
\rowcolor{TechBlue!10}
\textbf{HEAL}         & 32  &   384      & \textbf{49.34} & \textbf{40.00} & \textbf{44.67} & \underline{62.19} & \textbf{12.20} & \textbf{37.20} & \textbf{62.80}     & \textbf{19.76}    & \textbf{41.28} \\
\bottomrule
\end{tabular}
\caption{More performance comparison between our framework HEAL and other entropy regularization baselines evaluated on the Qwen3-1.7B-Base model under identical training dataset configurations.}
\label{tab:apx_details_df}
\end{table*}
\begin{figure*}[!t]    
    % \vspace{-5pt}
	\centering  
	\includegraphics[width=\linewidth]{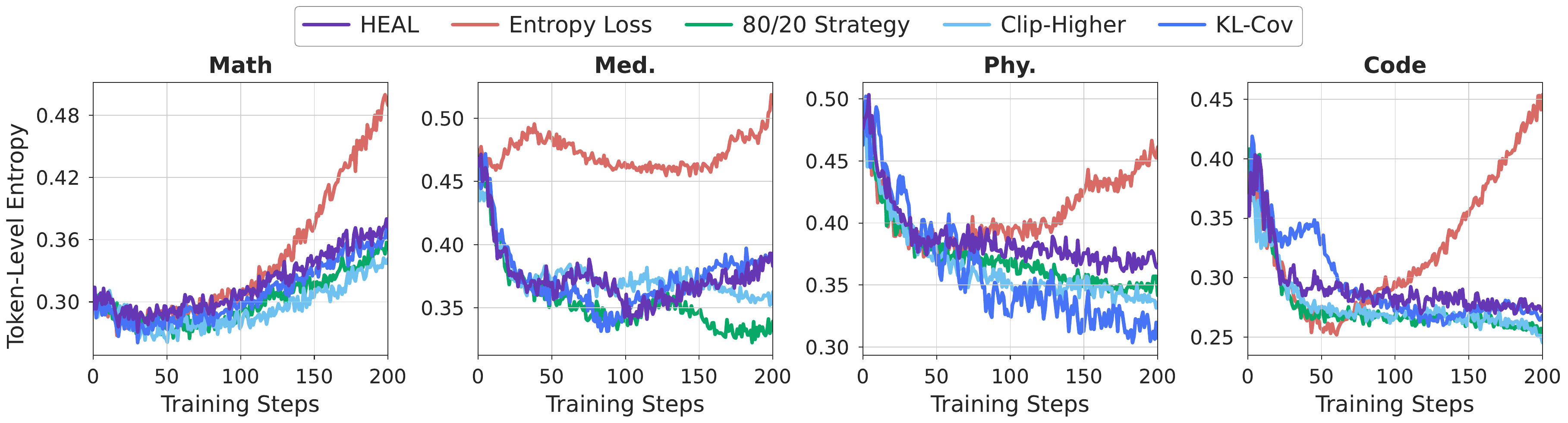}  
    % \vspace{-20pt}
	% \caption{The variation curves of average token entropy for the Qwen3-1.7B-Base model across three different target domains and under various entropy control methods.}
    \caption{Details of separate average token-level entropy curves on Math,  Medicine (Med.), Physics (Phy.), and Code domains. Evaluated on the Qwen3-1.7B-Base model under different entropy regularization methods and our framework HEAL.}
	\label{fig:more_details_en_aly}  
    % \vspace{-5pt}
\end{figure*}

\definecolor{blue}{HTML}{4402B6}
\usetikzlibrary{pgfplots.groupplots}
\usetikzlibrary{calc}

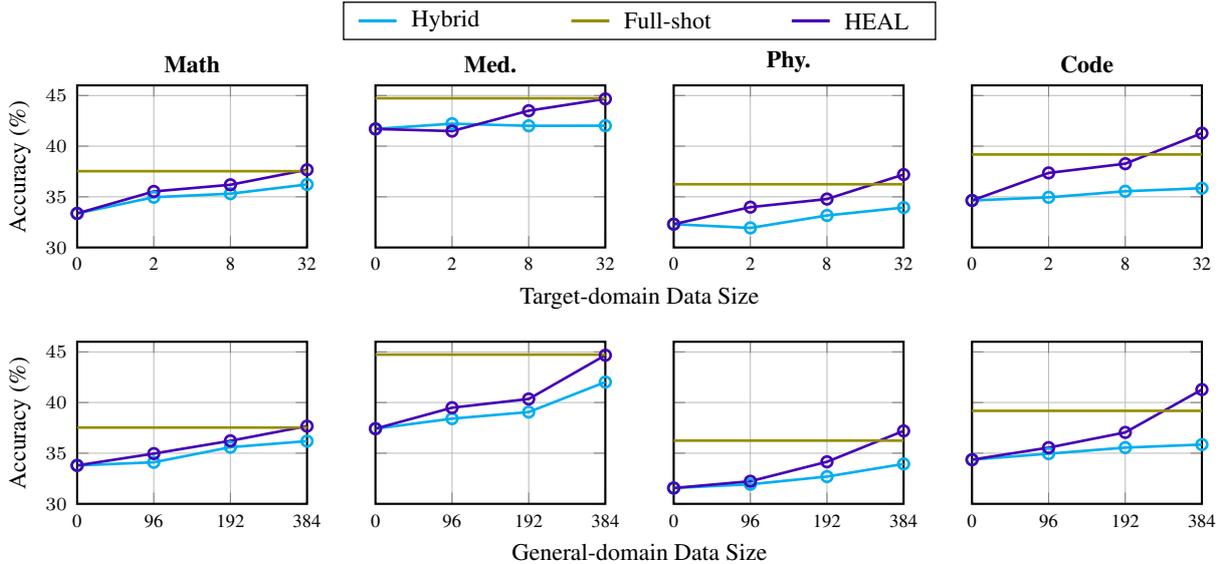
\begin{figure*}[t]
    \centering
    \hspace{-9pt}
    \begin{tikzpicture}[
        every axis/.append style={
            font=\small,
        }
    ]
    
    \begin{groupplot}[
        group style={
            group size=4 by 2,
            horizontal sep=0.9cm,
            vertical sep=1.25cm,
            group name=myplots,
            y descriptions at=edge left,
            x descriptions at=edge bottom,
        },
        width=0.19\textwidth,
        height=0.135\textwidth,
        grid=both,
        major grid style={line width=0.4pt, draw=gray!50},
        minor grid style={line width=0.25pt, draw=gray!20},
        tick label style={font=\scriptsize},
        label style={at={(axis description cs:0.25,0.5)},anchor=south, font=\small},
        axis lines=box,
        axis line style={black, thick},
        scale only axis,
        % 根据数据范围统一调整y轴
        ymin=30, ymax=46,
        xmin=0, xmax=3,
        xtick={0,1,2,3},
        title style={yshift=-1ex, font=\small\bfseries},
    ]
    
    % ========== 第一行子图 (Target-domain Scaling) ==========
    
    % (a) Domain A
    \nextgroupplot[title={Math}, ylabel={Accuracy (\%)}, xticklabels={0,2,8,32}]
    \addplot[color=cyan, mark=o, line width=1pt] coordinates {(0, 33.36) (1, 34.96) (2, 35.31) (3, 36.22)};
    \addplot[color=blue, mark=o, line width=1pt] coordinates {(0, 33.36) (1, 35.53) (2, 36.19) (3, 37.67)};
    \addplot[color=olive, line width=1pt] coordinates {(0, 37.53) (3, 37.53)};

    % (c) Domain C
    \nextgroupplot[title={Med.}, xticklabels={0,2,8,32}]
    \addplot[color=cyan, mark=o, line width=1pt] coordinates {(0, 41.70) (1, 42.22) (2, 42.01) (3, 42.02)};
    \addplot[color=blue, mark=o, line width=1pt] coordinates {(0, 41.70) (1, 41.49) (2, 43.50) (3, 44.67)};
    \addplot[color=olive, line width=1pt] coordinates {(0, 44.73) (3, 44.73)};

    % (d) Domain D
    \nextgroupplot[title={Phy.}, xticklabels={0,2,8,32}]
    \addplot[color=cyan, mark=o, line width=1pt] coordinates {(0, 32.30) (1, 31.93) (2, 33.16) (3, 33.94)};
    \addplot[color=blue, mark=o, line width=1pt] coordinates {(0, 32.30) (1, 33.98) (2, 34.78) (3, 37.20)};
    \addplot[color=olive, line width=1pt] coordinates {(0, 36.24) (3, 36.24)};

    % (b) Domain B
    \nextgroupplot[title={Code}, xticklabels={0,2,8,32}]
    \addplot[color=cyan, mark=o, line width=1pt] coordinates {(0, 34.64) (1, 34.95) (2, 35.55) (3, 35.85)};
    \addplot[color=blue, mark=o, line width=1pt] coordinates {(0, 34.64) (1, 37.36) (2, 38.27) (3, 41.28)};
    \addplot[color=olive, line width=1pt] coordinates {(0, 39.18) (3, 39.18)};
    
    % ========== 第二行子图 (General-domain Scaling) ==========
    
    % (e) Domain E
    \nextgroupplot[title={}, ylabel={Accuracy (\%)}, xticklabels={0,96,192,384}]
    \addplot[color=cyan, mark=o, line width=1pt] coordinates {(0, 33.79) (1, 34.10) (2, 35.59) (3, 36.19)};
    \addplot[color=blue, mark=o, line width=1pt] coordinates {(0, 33.79) (1, 34.96) (2, 36.22) (3, 37.67)};
    \addplot[color=olive, line width=1pt] coordinates {(0, 37.53) (3, 37.53)};
    
    % (f) Domain F

    % (g) Domain G
    \nextgroupplot[title={}, xticklabels={0,96,192,384}]
    \addplot[color=cyan, mark=o, line width=1pt] coordinates {(0, 37.42) (1, 38.41) (2, 39.06) (3, 42.02)};
    \addplot[color=blue, mark=o, line width=1pt] coordinates {(0, 37.42) (1, 39.49) (2, 40.35) (3, 44.67)};
    \addplot[color=olive, line width=1pt] coordinates {(0, 44.73) (3, 44.73)};

    % (h) Domain H
    \nextgroupplot[title={}, xticklabels={0,96,192,384}]
    \addplot[color=cyan, mark=o, line width=1pt] coordinates {(0, 31.56) (1, 31.93) (2, 32.69) (3, 33.94)};
    \addplot[color=blue, mark=o, line width=1pt] coordinates {(0, 31.56) (1, 32.23) (2, 34.15) (3, 37.20)};
    \addplot[color=olive, line width=1pt] coordinates {(0, 36.24) (3, 36.24)};

    \nextgroupplot[title={}, xticklabels={0,96,192,384}]
    \addplot[color=cyan, mark=o, line width=1pt] coordinates {(0, 34.36) (1, 34.95) (2, 35.55) (3, 35.85)};
    \addplot[color=blue, mark=o, line width=1pt] coordinates {(0, 34.36) (1, 35.55) (2, 37.05) (3, 41.28)};
    \addplot[color=olive, line width=1pt] coordinates {(0, 39.18) (3, 39.18)};
    
    \end{groupplot}
    
    % 公共图例
    \node[
        anchor=south, 
        draw=black,       % 【修改1】黑色边框
        line width=0.5pt, % 边框粗细
        fill=white,       % 白色填充
        inner ysep=1.5pt,  % 【关键】上下仅留 1.5pt 缝隙，使高度紧贴文本
        inner xsep=5pt,    % 左右留 5pt，保持美观
        yshift=0.3cm      % 向上微调位置
    ] at ($(myplots c1r1.north west)!0.5!(myplots c4r1.north east) + (0, 0.3cm)$) {
        \begin{tikzpicture}
            \draw[color=cyan, mark=o, line width=1.2pt] (0,0) -- (0.5,0) node[right, black, font=\small] {Hybrid};
            \draw[color=olive, line width=1.2pt] (2.8,0) -- (3.3,0) node[right, black, font=\small] {Full-shot}; 
            \draw[color=blue, mark=o, line width=1.2pt] (5.7,0) -- (6.2,0)  node[right, black, font=\small] {HEAL};
        \end{tikzpicture}
    };
    
    % 公共x轴标签
    \node[anchor=north, font=\small] at ($(myplots c1r1.south)!0.5!(myplots c4r1.south) + (0, -0.4cm)$) {Target-domain Data Size};
    \node[anchor=north, font=\small] at ($(myplots c1r2.south)!0.5!(myplots c4r2.south) + (0, -0.4cm)$) {General-domain Data Size};
    
    \end{tikzpicture}
    \caption{Comprehensive results of the Qwen3-1.7B-Base model across four target domains with respect to different data sizes: (\textit{Top}) Varying target-domain data size while keeping general-domain data fixed; (\textit{Bottom}) Varying general-domain data size while keeping target-domain data fixed. 
    }
    \label{fig:scaling_details}
\end{figure*}

\section{Additional Experiment Results}
\label{sec:appendix C}
\subsection{Entropy Details of Main Results}
To provide a more granular perspective on the results presented in Figure~\ref{fig:preliminary_entropy_curve}, we detail the entropy curves across all four target domains, including the Math domain, in Figure~\ref{fig:more_details_en_main}. Our empirical observations across these diverse domains strongly support the hypothesis that in vanilla low-resource RLVR scenarios, models are highly susceptible to entropy collapse. 
While standard hybrid training (Hybrid) mitigates this collapse to some extent, the results in Figure~\ref{fig:more_details_en_main} demonstrate the superior ``\textit{healing}'' capability of our HEAL framework. In all four domains, HEAL consistently restores entropy from a collapsed state to levels comparable to the Full-shot baseline. This consistent effectiveness underscores the robust generalization capabilities and stability of the HEAL framework across varying target domains.

\subsection{Additional Results on the Math Domain}
\label{apx:more_math}
To further validate the robustness and architectural agnosticism of HEAL within the Math domain, we extended our experiments to encompass various Qwen3 parameter scales and the LLaMA3 architecture. As reported in Table~\ref{tab:math_only}, the results are highly consistent with our primary findings:
First, HEAL significantly outperforms the Hybrid baseline across all evaluated model scales and architectures. Second, HEAL achieves performance that is competitive with, or even superior to, the Full-shot upper bound on several benchmarks. Third, the Only-General baseline often yields results inferior to Few-shot baselines, further highlighting the unique value of HEAL in successfully aligning cross-domain exploration diversity.

% \input{table/Appendix_C/MainResult_0.6B}

% \subsection{Performance on Other Models}
% \label{apx:more_models}
% To investigate whether the effectiveness of HEAL is sensitive to model parameter size, we conducted experiments using Qwen3-0.6B-Base (0.6B) and Qwen3-8B-Base (8B). For the 0.6B model, we performed evaluations across all four target domains; for the 8B model, due to computational constraints, we focused our validation on the Math domain. 
% The experimental results, summarized in Table~\ref{tab:math_qwen3} and Table~\ref{tab:detailed_results_0.6b}, confirm that HEAL consistently exceeds the Hybrid baseline and approaches Full-shot performance regardless of model capacity. These findings support the conclusion that HEAL is a scale-agnostic optimization framework, enabling efficient exploration strategies under data-scarce conditions for models of varying sizes. 

\subsection{More Results of Entropy Regularization Methods}
\label{apx:more_entropy_methods}
This subsection provides a detailed analysis of the performance on individual benchmarks across all target domains using various methods of entropy regularization. Results in Table~\ref{tab:apx_details_df} show that the standard Entropy Loss baseline leads to poor performance on many tasks. As illustrated in Figure~\ref{fig:more_details_en_aly}, maximizing entropy with a simple scalar-based regularization often causes an entropy explosion, an unintended effect where the model produces disordered and incoherent text due to excessively high entropy,
Furthermore, while other competitive methods for entropy regularization keep entropy at moderate levels, their final task performance is still much lower than HEAL. This observation reveals an important insight: high performance in RLVR requires more than just keeping entropy at a certain level; it depends on the dynamics of entropy over time. By aligning the entropy dynamic patterns of the target domain with those of the general-domain, HEAL ensures the model learns when to explore and when to focus on specific logical paths.

\subsection{Details about Impact of Data Size}
\label{apx:more_details_of_size}

Figure~\ref{fig:scaling_details} details  how performance changes as the amount of data increases across four distinct domains (Math, Code, Medicine, and Physics). Consistent with the analysis in Section~\ref{subsec:scaling}, HEAL exhibits a significantly steeper growth rate than the Hybrid baseline, whether increasing target domain (top row) or general-domain (bottom row) data sizes. Taking the Code and Medicine domains as prime examples, HEAL effectively unlocks the potential of the data, rapidly diverging from the Hybrid and even surpassing the Full-shot upper-bound at specific data scales (e.g., when target domain data size reaches 32).

\section{Further Analysis}
\label{sec:appendix D}

\begin{figure*}[!t]    
    % \vspace{-5pt}
	\centering  
	\includegraphics[width=\linewidth]{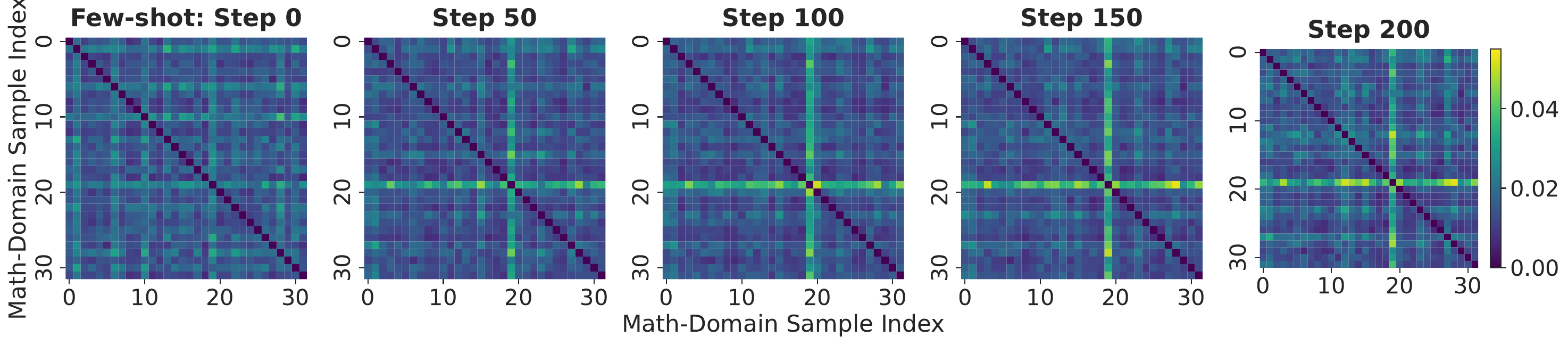} 
    \includegraphics[width=\linewidth]{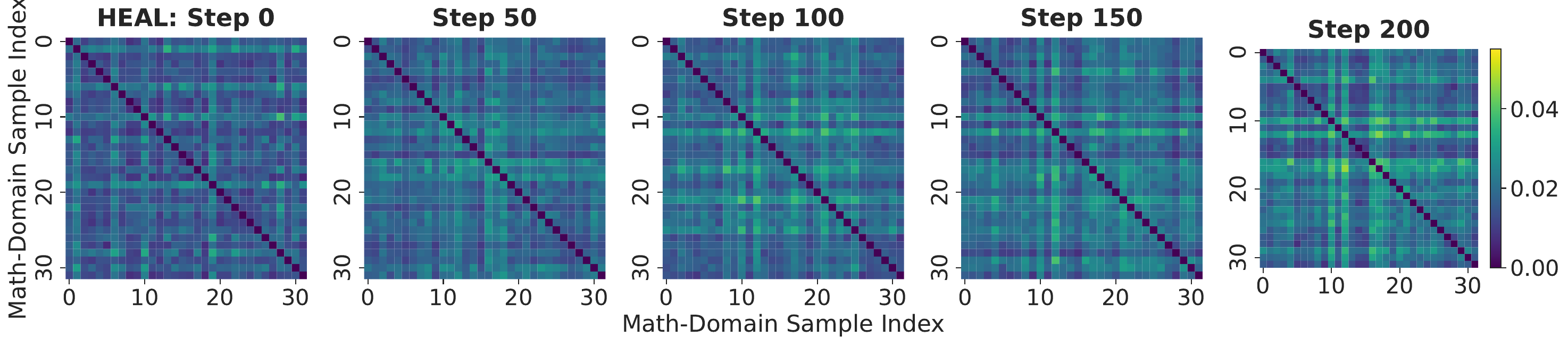} 
    % \vspace{-20pt}
	% \caption{The variation curves of average token entropy for the Qwen3-1.7B-Base model across three different target domains and under various entropy control methods.}
    \caption{Visualization of Entropy Dynamics (ED) diversity evolution. The heatmaps display the pairwise distances between EDs of generated samples in the Math domain across training steps 0--200. (\textit{Top}): The Few-shot baseline maintains consistently low inter-sample distances (predominantly dark regions), indicating a high degree of homogeneity and limited exploration. (\textit{Bottom}): HEAL exhibits a progressive and widespread increase in distances (shifting to brighter regions), demonstrating that our framework effectively diversifies entropy trajectories.}
	\label{fig:hitmaps}  
    % \vspace{-5pt}
\end{figure*}
\subsection{Exploration Diversity}

To investigate the exploration mechanism underlying HEAL, we analyze the model's Entropy Dynamics (EDs) diversity. Mathematically defined in Section~\ref{sec:reward}, each ED vector captures the evolution of uncertainty throughout the reasoning process. We hypothesize that the diversity of ED patterns serves as a proxy for the diversity of reasoning paths; specifically, a larger distributional distance between ED sequences indicates that the model is exploring a more heterogeneous set of generation strategies rather than collapsing into a single exploration pattern.

The distance between two EDs is defined as $d = -s_{\text{Ours}}$, based on the similarity metric in Equation~\ref{eq:ourds}.
Figure~\ref{fig:hitmaps} visualizes the distances between the EDs of generated samples in a target domain (e.g Math) throughout training. Comparing the baseline vanilla Few-shot RLVR baseline (top) with our HEAL framework (bottom), we observe distinct evolutionary behaviors. 
Under the Few-shot setting, the distances among EDs remain consistently low, indicating a high degree of homogeneity. The dominance of low-distance regions suggests that the model converges to a narrow set of ED patterns, reflecting limited exploration and a tendency toward diversity collapse. 
In contrast, HEAL exhibits a progressive increase in distances among EDs. This trend indicates that the model under HEAL effectively diversifies its entropy trajectories, thereby mitigating the tendency to converge to few reasoning patterns.
This observation validates the effectiveness of our EDA reward. By incorporating soft reward signals derived from general-domain EDs, HEAL encourages the model to maintain higher entropy, effectively unlocking the data's potential by preventing premature convergence to local optima.

\definecolor{blue}{HTML}{4402B6}
\usetikzlibrary{pgfplots.groupplots}
\begin{figure}[t]
    % \hspace*{-0.3cm}
    % \hspace*{0.1cm}
    \vspace{-6pt}
    \centering
    \begin{tikzpicture}
        \begin{groupplot}[
            group style={
                group size=2 by 1, % 一行两列
                horizontal sep=0.6cm, % 子图之间的水平间距
                vertical sep=1.25cm, % 子图之间的垂直间距
                group name=myplots, % 为组命名，便于后续引用
            },
            width=0.26\textwidth, % 每个子图的宽度
            height=0.28\textwidth, % 每个子图的高度
            grid=both, % 显示网格
            major grid style={line width=0.4pt, draw=gray!50},
            minor grid style={line width=0.25pt, draw=gray!20},
            tick label style={font=\small},
            label style={at={(axis description cs:0.25,0.5)},anchor=south, font=\small},
            legend cell align=left,
            title style={yshift=-1ex, font=\small},
            % 统一的轴设置
            axis lines=box, % 完整的方框边界
            axis line style={black, thick}, % 坐标轴线条样式
        ]
        % 第一个子图
        \nextgroupplot[
            ylabel={Pass@$k$ Result},
            xmin=0, xmax=2,
            ymin=1, ymax=28,
            xtick={0,1,2},
            xticklabels={1,5,10},
            % 调整y轴标签位置，使其紧贴刻度
            % ylabel style={at={(ticklabel* cs:0.5)}, xshift=-0.8cm, anchor=center, rotate=90},
            % legend pos=north west, % 暂时保留图例位置，稍后我们将移动它
            legend style={
                at={(1.05,-0.35)},
                anchor=south,
                legend columns=3,
                draw=none,
                font=\small,
                /tikz/column sep=3pt,
            },
            title={(a) Qwen3-1.7B-Base} % 子图标题
        ]
            \addplot[
                color=olive,
                mark=o,
                line width=1pt
            ] coordinates {
                (0, 8.32) 
                (1, 14.65) 
                (2, 17.37)
            };
            \addlegendentry{Full-shot}
            \addplot[
                color=cyan,
                mark=o,
                line width=1pt
            ] coordinates {
                (0, 3.41) 
                (1, 8.64) 
                (2, 12.57)
            };
            \addlegendentry{Few-shot}
            \addplot[
                color=blue,
                mark=o,
                line width=1pt
            ] coordinates {
                (0, 8.97) 
                (1, 16.55) 
                (2, 19.76)
            };
            \addlegendentry{HEAL}
        
        % 第二个子图
        \nextgroupplot[
            xmin=0, xmax=2,
            ymin=1, ymax=28,
            xtick={0,1,2},
            xticklabels={1,5,10},
            % ytick pos=right, % 将y轴刻度放在右侧
            % yticklabel style={
            %     font=\small,
            %     xshift=0pt, % 稍微向右偏移，避免与网格线重叠
            % },
            yticklabels={},
            axis lines=box, % 完整的方框边界
            title={(b) Qwen3-4B-Base} % 子图标题
        ]
            \addplot[
                color=cyan,
                mark=o,
                line width=1pt
            ] coordinates {
                (0, 16.17) 
                (1, 21.23) 
                (2, 22.75)
            };
            % 不添加图例，使用统一的图例
            \addplot[
                color=blue,
                mark=o,
                line width=1pt
            ] coordinates {
                (0, 17.42) 
                (1, 23.86) 
                (2, 26.95)
            };
            \addplot[
                color=olive,
                mark=o,
                line width=1pt
            ] coordinates {
                (0, 11.38) 
                (1, 19.67) 
                (2, 23.95)
            };
        \end{groupplot}

    \end{tikzpicture}
    
    \caption{Pass@$k$ results on LiveCodeBench v5 for Qwen3 models. The performance of the 1.7B- and 4B-Base models is compared across Few-shot, Full-shot, and HEAL settings. We investigate the impact of Pass@$k$ by varying $k$ among 1, 5, and 10.}
    \label{fig:passk}
    % \vspace{-10pt}
\end{figure}
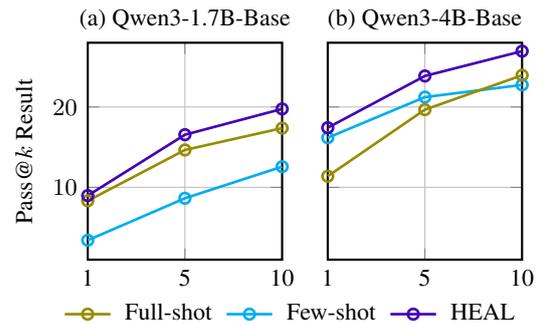
\subsection{Evaluating Pass@k Metrics on LiveCodeBench}
\label{apx:passk}

To further evaluate the exploratory potential of HEAL from a discovery perspective, we employ the Pass@$k$ metric on the LiveCodeBench benchmark. As shown in Figure~\ref{fig:passk}, we compared the performance of Qwen3-1.7B-Base and Qwen3-4B-Base models under Few‑shot, Full‑shot, and HEAL settings, with $k$ varied among {1, 5, 10}. A key observation on the Qwen3-4B-Base model is that while the Few-shot baseline performs reasonably at $k=1$, it lags significantly behind the Full-shot baseline at $k=10$. This gap suggests that Few-shot models often suffer from \textit{exploration collapse}, where the model over-exploits a few high-probability paths and fails to explore alternative correct solutions.
In contrast, HEAL achieves superior performance across all values of $k$. It maintains high precision at low $k$ while demonstrating an exploratory breadth at high $k$ that rivals or exceeds the Full-shot baseline. This evidence suggests that HEAL successfully demonstrates a superior balance between exploration and exploitation, allowing the model to not only identify the most likely answer but also to maintain the potential to discover deeper reasoning paths in low-resource settings.

\end{document}